\documentclass{article}
\usepackage{iclr2027_conference,times}
\usepackage[T1]{fontenc}
\usepackage{amsmath,amssymb,booktabs,tabularx}
\usepackage{microtype}
\usepackage{graphicx}
\usepackage[percent]{overpic}
\newsavebox{\panelbox}
\newcommand{\panelmarktop}{9pt}
\newcommand{\panelmark}[2]{%
  \llap{\hbox to \wd\panelbox{\hspace{#1}\hspace{1pt}%
    \raisebox{\dimexpr\ht\panelbox-\panelmarktop\relax}[0pt][0pt]{%
      {\sffamily\bfseries\fontsize{9}{11}\selectfont #2}}\hss}}%
}
\newcommand{\panelgraphic}[3][0pt]{%
  \leavevmode
  \sbox{\panelbox}{#3}%
  \usebox{\panelbox}\panelmark{#1}{#2}%
}
\newcommand{\threepanelgraphic}[2][0pt]{%
  \leavevmode
  \sbox{\panelbox}{#2}%
  \usebox{\panelbox}%
  \panelmark{#1}{a}%
  \panelmark{\dimexpr\wd\panelbox/3+#1\relax}{b}%
  \panelmark{\dimexpr\wd\panelbox*2/3+#1\relax}{c}%
}

\usepackage{xcolor}
\usepackage[breakable]{tcolorbox}
\usepackage{hyperref}
\usepackage{url}

\title{Counting on Thinking: Tracing Evidence Integration in Language Models}
\author{Jingming Xue$^{1}$, Robert C. Wilson$^{1}$, and Hua-dong Xiong$^{1,2}$ \\
$^1$Georgia Institute of Technology, Atlanta, GA \\
$^2$University of Chicago, Chicago, IL \\
\texttt{jxue93@gatech.edu} \\
\texttt{rwilson337@gatech.edu} \\
\texttt{hdx@gatech.edu}
}

\iclrfinalcopy

\begin{document}
\maketitle
\fancyhead{}

\begin{abstract}
Computational resources are finite, so no intelligent system can afford extended computation for every decision. Humans and animals handle frequent operations with automatic \textit{System~1} processes and reserve costly \textit{System~2} computation for problems that warrant it. Large language models (LLMs) can likewise spend extra inference-time computation on hard problems, yet their direct answers struggle even with counting, an elementary integration of input that humans and animals perform automatically. We ask why such an operation requires thinking in LLMs. Evidence integration is so basic that psychology and neuroscience have long used it to probe decision-making; our LLM version presents one letter per conversational turn and asks the model which of two target letters appeared more often. A running difference between the two counts solves the task optimally, weighting every letter equally, and fits this format exactly: the tokens at each turn can represent the difference accumulated over all earlier letters and update it with the new one. Direct responses nonetheless weighted evidence unevenly, with a strong recency effect, and assigned less probability to the correct answer as the task grew harder. Thinking improved performance and made the integration weights nearly uniform, yet final-query attention stayed concentrated on the ends of the sequence in both modes. Instead, reasoning trajectories showed the model revisiting the input, recounting the letters, and checking intermediate counts that the answer then read, suggesting that thinking constructs the accumulated count that direct responses lack rather than reading out one already formed. This recounting cost reasoning tokens that grew with the number of letters far more than with coherence. Nor did outcome feedback bring the computation into direct responses: under in-context reinforcement learning (ICRL), performance deteriorated over repeated games and the recency effect strengthened, yet the models grew more confident. Humans and animals amortize such elementary computations into automatic processes, whereas current LLMs still pay for them with thinking on every trial. Which operations learning can make directly available is a central question for how future models allocate computation.
\end{abstract}

\section{Introduction}
\label{sec:introduction}

Computation is a limited resource, and every intelligent system must decide where to spend it. Humans and animals handle most decisions with fast, automatic processes and recruit slower, effortful computation only when a situation demands it. The distinction between \textit{System~1} and \textit{System~2} captures this division \citep{evans_dualprocess_2013}, and resource-rational accounts explain its logic: effortful computation pays off only when it improves decisions enough to justify its cost \citep{shenhav_expected_2013,griffiths_rational_2015,lieder_resourcerational_2020}, whereas computations required often can be amortized, so that their cost is paid largely during learning \citep{gershman_amortized_2014}. Integrating incoming evidence is one such computation. Performed reliably by a fast pathway, it leaves effortful computation free for the problems that need it.

Large language models (LLMs) can likewise spend additional computation at inference time, and we treat their non-thinking and thinking modes as functional analogues of \textit{System~1} and \textit{System~2}. Their \textit{System~1} has conspicuous gaps. LLMs solve complex problems \citep{openai_openai_2024,guo_deepseekr1_2025}, yet their direct answers have struggled with elementary letter counting, as in the widely noted failure to count the occurrences of ``r'' in ``strawberry'' \citep{fu_when_2026}. Thinking lets models solve harder problems: generating intermediate steps helps on arithmetic and symbolic tasks that models cannot answer directly \citep{wei_chainofthought_2022,nye_show_2021,lanchantin_learning_2023} and extends what a fixed transformer can compute \citep{merrill_expressive_2024,li_chain_2024}. The gain has a price: thinking can spend substantial computation on simple questions with little additional benefit \citep{chen_not_2025}. Counting letters that arrive one at a time is the textual analogue of integrating sensory evidence, and a single counter, updated once per turn, solves it exactly. Theoretical work shows how transformers can implement counting \citep{weiss_thinking_2021} and examines how this ability depends on model capacity \citep{yehudai_when_2024}. We therefore ask to what extent this simple integration fails in direct responses, where the failure lies, and what thinking adds when every letter is already in the context. We consider two hypotheses. The construction hypothesis holds that thinking forms a count that direct responses lack, converting letters scattered across the input into an accumulated state through additional steps. The readout hypothesis instead assumes that the count already exists and that thinking improves how the answer uses it. The distinction matters because it locates the failure of direct responses either in forming the count or in using it, which determines what a fast pathway would have to learn.

Accumulating evidence over time is so basic to how humans and animals guide action that psychology and neuroscience have long studied it as a model of decision-making \citep{gold_neural_2007,bogacz_physics_2006,brunton_rats_2013}. Behavioral integration kernels estimate from choices how strongly evidence at each moment influences the decision \citep{kiani_bounded_2008,keung_divisive_2020,xiong_humans_2025}, and the optimal kernel depends on the task: discounting older evidence is adaptive in changing environments \citep{glaze_normative_2015}, whereas counting across a whole sequence is optimal only with uniform integration weights, which weight every item equally (Appendix~\ref{sec:kernel-fitting}). LLMs are known to use context unevenly, with recency bias in few-shot prompting \citep{zhao_calibrate_2021} and weaker use of the middle of long contexts \citep{liu_lost_2024}. For counting, the kernel turns this unevenness into a direct test: any deviation from uniform weights marks a departure from the rule. The kernel is a functional description of how input evidence maps to the answer under the tested protocol, not an estimate of an internal accumulator. Simplicity is what makes this test possible: the fully specified counting rule lets us characterize how thinking changes temporal evidence weighting, beyond measuring whether it improves the final answer.

We apply this framework directly to LLMs: models observe letters one per conversational turn and choose which of two target letters occurred more often while ignoring distractors (Figure~\ref{fig:base-icl-dialogue}). Across seven models and a range of sequence lengths, coherences, and distractor densities, we compare direct answers with answers generated after thinking from an identical history, using integration kernels to measure how evidence at each position shapes the answer and final-query attention to record where the answer query attends at its last step. Because the kernel reflects the entire computation and final-query attention only its last step, we examine whether more uniform evidence use accompanies more uniform access to the letters at the answer.

Even on this task, direct responses depart from the rule in a structured way. They weight evidence unevenly: their integration weights skew toward the end of the sequence, a recency effect, and stay low in its middle, a lost-in-the-middle profile. Their correct-answer negative log-likelihood also rises on longer sequences. Thinking improves performance and makes the integration weights more uniform, approaching the equal weighting the rule requires, at a reasoning-token cost that grows with the number of letters far more than with coherence. Final-query attention in both modes concentrates on the ends of the sequence, even as thinking makes evidence use more uniform. Direct answers attend most to the latest letter yet do not reliably reflect the accumulated evidence. Selected reasoning trajectories illustrate how thinking processes the evidence. The model revisits the letters and writes out intermediate counts that the final query attends to, regenerating information already present in its input. These observations suggest that thinking constructs explicit intermediate counts.

LLMs adapt to new tasks through statistical learning in context, reorganizing their internal representations to fit the task structure \citep{brown_language_2020,wei_larger_2023,park_iclr_2025,xiong_large_2026}. Outcome feedback across repeated games might therefore teach the model to integrate the letters in its residual stream, making the integration kernels of direct responses progressively more uniform. We test this with in-context reinforcement learning (ICRL): models play ten games in one conversation with correctness feedback after each decision \citep{laskin_incontext_2022,monea_llms_2024}. As games accumulate, correct-answer negative log-likelihood rises and integration kernels show a stronger recency effect, yet the models become more confident on average. The policy thus becomes more concentrated while continuing to favor recent evidence. Where humans and animals amortize elementary computations into fast processes, current LLMs still recover counting through thinking, spending computation and context on an operation that a fast pathway could perform.

This work makes four contributions. We introduce a sequential counting protocol whose reference computation is fully specified, so that any departure from uniform integration weights is interpretable. Behavioral integration kernels characterize how thinking changes the temporal distribution of evidence weights, from the strong recency of direct responses toward the uniform profile specified by the rule. Comparing kernels with final-query attention dissociates evidence use from evidence access at the answer: thinking makes evidence use nearly uniform while measured final-query attention to the original letters remains concentrated on the sequence ends. Finally, an ICRL test shows that ten games with outcome feedback do not make direct evidence weighting more uniform.

\section{Methods}
\label{sec:methods}

\subsection{Sequential letter-counting task}
\label{sec:task}
\begin{figure}[htbp]
\centering
\includegraphics[width=\textwidth]{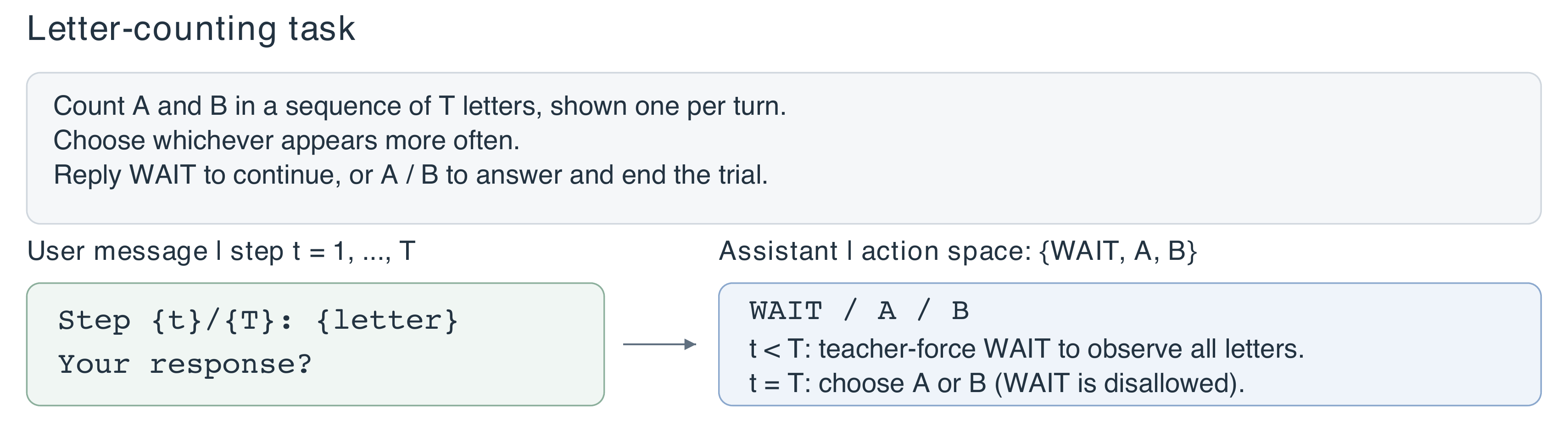}
\caption{\textbf{The sequential letter-counting task.} A $T$-step trial with example target letters A and B. Letters arrive one per turn, and the model responds with \texttt{WAIT}, A, or B; at the final step, \texttt{WAIT} is disallowed. The two experiments differ in how responses are controlled. In the single-game experiment, each conversation contains one game: \texttt{WAIT} is teacher-forced at nonfinal steps so that the model observes the whole sequence, and at the final letter it answers either directly (non-thinking) or after thinking. In the ICRL experiment, each conversation contains ten non-thinking games: the model chooses its own responses, may answer before the final letter, and learns after each game only whether its answer was correct. The schematic depicts response control in the single-game experiment; ICRL games use the same message format. System instructions are shown schematically; Appendix~\ref{sec:prompts} gives the complete prompts.}
\label{fig:base-icl-dialogue}
\end{figure}

We asked models to identify which of two target letters occurred more often in a sequence presented one letter per conversational turn (Figure~\ref{fig:base-icl-dialogue}). The instructions specified the two targets and asked the model to count them while ignoring all other letters, which served as distractors. Each game used a new target pair and sequence.

For a sequence $\ell_1,\ldots,\ell_T$ with targets A and B, let $x_i=+1$ if $\ell_i$ is A, $x_i=-1$ if it is B, and $x_i=0$ for any other letter. A signed count then gives the answer:
\begin{equation}
z_0=0,\qquad z_i=z_{i-1}+x_i,\qquad a^\star=\operatorname{sign}(z_T),
\label{eq:reference-count}
\end{equation}
with $a^\star=+1$ denoting A. Each occurrence of a target letter thus contributes the same unit weight, independent of its position in the sequence, so the optimal integration weights are uniform. This recurrence defines the optimal case of the task; it does not assume that the model internally implements a running count or any particular accumulation algorithm.

\subsection{Experimental design}
\label{sec:base-design}
\label{sec:icl-design}
The single-game experiment asked what thinking adds once all the evidence has been received. Each sequence was presented in a fresh conversation, and the response at every nonfinal step was fixed to \texttt{WAIT}, so the model observed the whole sequence. At the final letter, each trial branched into a direct answer (non-thinking) and an answer produced after a reasoning continuation (thinking), both from an identical history. We varied one factor at a time around a baseline of 40 letters, coherence 0.10, and no distractors. Specifically, we varied sequence length (20 to 60 letters); coherence, the requested target-count difference relative to the number of target letters (0.05 to 0.20); and distractor density (0 or 0.20). Because coherence was fixed across lengths, longer sequences carried larger absolute count differences (Table~\ref{tab:conditions}).
The repeated-game experiment asked whether experience improves direct answers. Models played ten non-thinking games of 20 letters in one conversation and could wait for further evidence or answer early. After each decision, feedback stated only whether it was correct, while earlier observations, choices, and feedback remained in context. We refer to this as the in-context reinforcement learning (ICRL) experiment \citep{laskin_incontext_2022,monea_llms_2024}. Answers were scored by counting the two target letters across all 20 positions, in conditions with or without distractors. Complete conditions, sequence generation, and prompts are given in Appendix~\ref{sec:suppmethods}.

\subsection{Models}
We studied seven models: Qwen3.5-4B and Qwen3.5-9B \citep{qwen_qwen35_2026}, Gemma-4-E2B-it and Gemma-4-E4B-it \citep{gemma_2026}, Qwen3.6-35B-A3B \citep{qwen_qwen35_2026}, Inkling-Small \citep{lab_inkling_2025}, and Nemotron-3.5-Lightning-30B-A3B-BF16 \citep{nvidia_nvidia_2025}. The four Qwen3.5 and Gemma-4 models provided complete action scores and form the attention cohort; the other three returned top-20 token scores and contribute to the seven-model behavioral averages. Because the Qwen3.5 and Qwen3.6 models used mode-specific sampling settings, their mode contrast also includes a change in sampling. Sample sizes, scoring, and recorded layers are specified in Appendix~\ref{sec:cohort-conditions}.

\subsection{Measures}
\label{sec:measures}
We measured decision quality as the negative log-likelihood (NLL) of the correct answer, normalizing action scores over the legal actions at unit temperature. In the ICRL experiment we also report action entropy, computed from the executed sampling distribution, which measures how concentrated, and in this sense how confident, the policy is. Integration kernels regress, for each model, condition, and mode, the final score difference between the two target letters, $y_t$, on the signed stimulus codes of trial $t$,
\begin{equation}
y_t=b+\sum_{i=1}^{T}w_i x_{ti}+\varepsilon_t,
\end{equation}
by ridge regression (Appendix~\ref{sec:kernel-fitting}). The integration weight $w_i$ measures how strongly a target letter at position $i$ moves the final preference toward that letter. Kernels keep raw score scales, so their height reflects overall sensitivity to evidence and their profile its temporal distribution; under Eq.~\ref{eq:reference-count}, the optimal profile is uniform, weighting every position equally. Final-query attention is the attention that the query producing the answer assigns to the original letter tokens. Because individual heads specialize in function \citep{olsson_incontext_2022}, we used stimulus-selective heads chosen on separate conversation blocks and shared across single-game modes, normalizing their attention over the letters (Appendix~\ref{sec:attention-details}). Reasoning length is the number of generated reasoning tokens. Behavioral measures average all seven models and attention measures the four complete-score models. Single-game kernels use trials eligible in both modes; ICRL kernels use only games that observed all 20 letters, whereas ICRL NLL and action entropy also include early answers (Appendix~\ref{sec:analysis-details}).

\section{Results}
\label{sec:results}

\subsection{Thinking makes the uneven integration weights of direct responses nearly uniform}
\label{sec:results-direct}
\label{sec:results-thinking}
\begin{figure}[!htbp]
\centering
\renewcommand{\panelmarktop}{\dimexpr .01636\linewidth-2pt\relax}
\renewcommand{\threepanelgraphic}[2][0pt]{%
  \leavevmode\sbox{\panelbox}{#2}\usebox{\panelbox}%
  \panelmark{\dimexpr .10\wd\panelbox-10pt\relax}{a}%
  \panelmark{\dimexpr .40418\wd\panelbox-10pt\relax}{b}%
  \panelmark{\dimexpr .70835\wd\panelbox-10pt\relax}{c}%
}
\threepanelgraphic{\includegraphics[width=\linewidth]{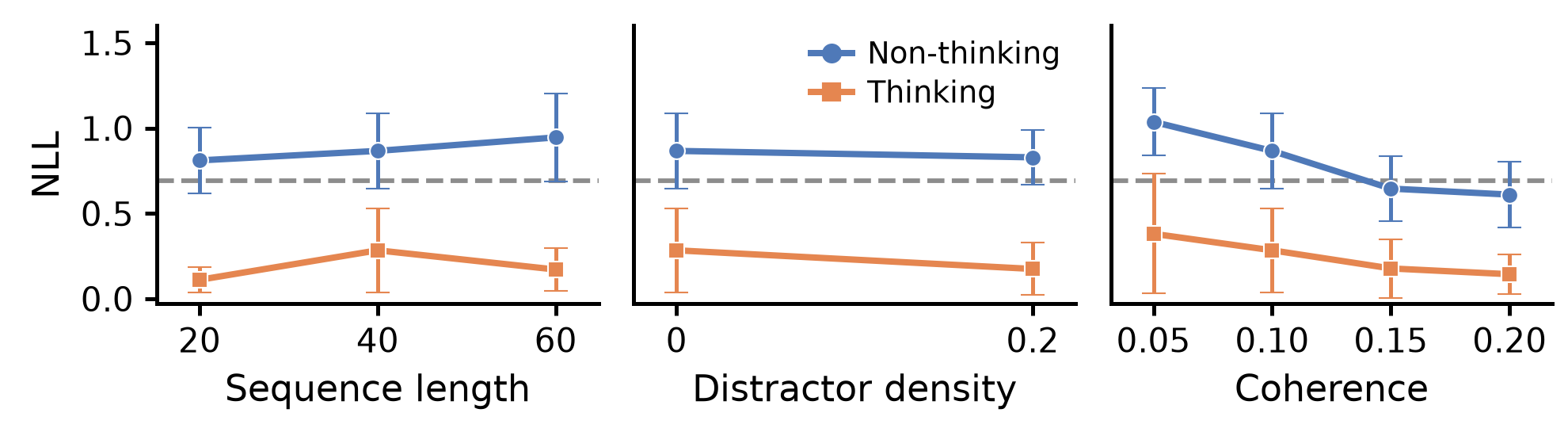}}
\par\vspace{-4pt}
\renewcommand{\panelmarktop}{\dimexpr .06\linewidth-2pt\relax}
\begin{minipage}[t]{.333\linewidth}
\panelgraphic[\dimexpr .34\linewidth-10pt\relax]{d}{\includegraphics[width=\linewidth]{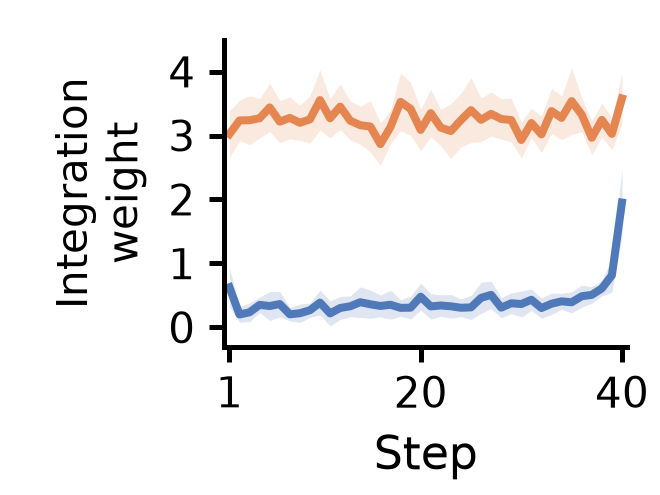}}
\end{minipage}%
\begin{minipage}[t]{.333\linewidth}
\panelgraphic[\dimexpr .34\linewidth-10pt\relax]{e}{\includegraphics[width=\linewidth]{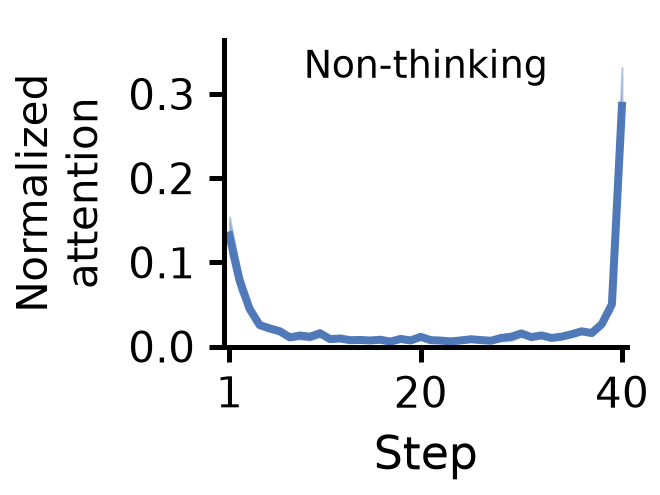}}
\end{minipage}%
\begin{minipage}[t]{.333\linewidth}
\panelgraphic[\dimexpr .34\linewidth-10pt\relax]{f}{\includegraphics[width=\linewidth]{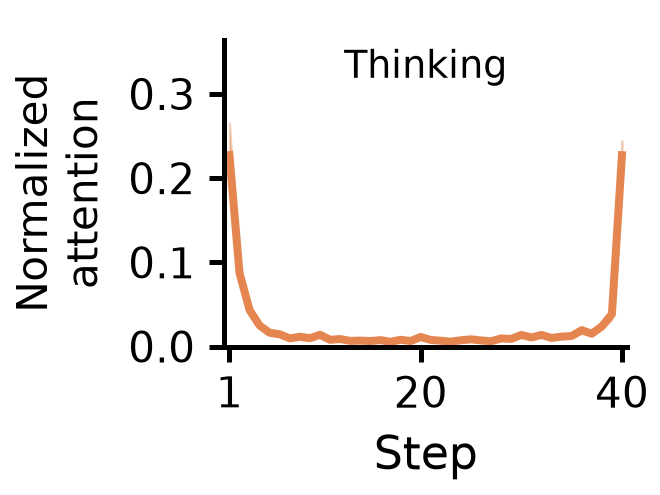}}
\end{minipage}
\caption{\textbf{Thinking lowers correct-answer NLL and makes integration weights nearly uniform, but final-query attention stays concentrated on the sequence ends.} (a--c) Correct-answer NLL by sequence length, distractor density, and coherence; dashed line: uninformed guess ($\ln 2$). (d) Integration kernels, fitted on the same trials in both modes. (e,f) Final-query stimulus attention from stimulus-selective heads in non-thinking (e) and thinking (f) modes, normalized over the original letters within each trial so that each curve sums to one; its scale is not comparable to the kernel in d. Panels d--f use 40 letters, coherence 0.10, and no distractors. Panels a--d average seven models and e,f the four complete-score models. Blue: non-thinking; orange: thinking. Means $\pm$ model SEM ($n=7$ models in a--d; $n=4$ in e,f). Estimator details in Appendices~\ref{sec:kernel-fitting} and~\ref{sec:attention-details}.}
\label{fig:base-performance}
\end{figure}

Direct responses performed poorly on a task that only asks which of two letters appeared more often. Without thinking, the seven-model mean of correct-answer NLL stayed near the level of an uninformed guess and exceeded it in all but the two highest coherence conditions (Figure~\ref{fig:base-performance}a--c; error bars show model SEM). The task became harder as sequences grew longer, even though the count differences were larger. Lower coherence also made the task harder. NLL reflected both effects: it rose with sequence length and fell as coherence increased. Adding distractors did not increase NLL.

Integration kernels locate the departure from the counting rule, which weights every position equally. Non-thinking integration weights were instead uneven, with a small initial peak (primacy), little weight through most of the sequence, and a strong skew toward its end (recency; Figure~\ref{fig:base-performance}d; other conditions in Appendix Figure~\ref{fig:base-condition-kernels}). Direct answers thus depended disproportionately on the final letters.

Thinking lowered correct-answer NLL at every length and coherence level (Figure~\ref{fig:base-performance}a,c) and changed how evidence reached the answer. Thinking kernels were nearly uniform across positions, with the recency effect of direct responses largely gone, and larger overall (Figure~\ref{fig:base-performance}d; individual-model kernels in Appendix Figure~\ref{fig:individual-base-kernel}). Because kernel height scales with the range of the action scores, the more decisive thinking answers contribute to this increase, so the comparison rests on the profile rather than on its height. Because both modes branched from the same sequence and history, this change arose in what the model did between the last letter and its answer.

\subsection{Thinking revisits the evidence and generates intermediate counts}
\label{sec:results-attention}
\label{sec:results-cost}
The simplest account of what thinking adds is more even evidence access, with the final answer attending evenly to the original letters. In direct responses, final-query attention peaked weakly at the first letter and strongly at the last (Figure~\ref{fig:base-performance}e), a profile whose ends match the primacy and recency of the non-thinking kernel, although the first-position peak may partly reflect the general tendency of LLMs to attend to early tokens \citep{barbero_why_2025}. The direct answer thus attended most to the latest letter, whose turn could already have held the signed count of Eq.~\ref{eq:reference-count} accumulated over earlier turns, yet had correct-answer NLL near the uniform-guess reference. Thinking left a similar pattern, with the same stimulus-selective heads: attention still concentrated on the first and last letters and assigned little to the middle (Figure~\ref{fig:base-performance}f; other lengths in Appendix Figure~\ref{fig:base-attention-lengths}), even though the behavioral weights had become more uniform. More uniform evidence use thus accompanied persistently uneven final-query attention to the original letters. This rules out more even access to the letters at the answer, but not readout, because reading out a count formed during observation would also draw attention to the latest letter; the reasoning trajectories bear on that distinction.

\begin{figure}[!htbp]
\centering
\includegraphics[width=\linewidth]{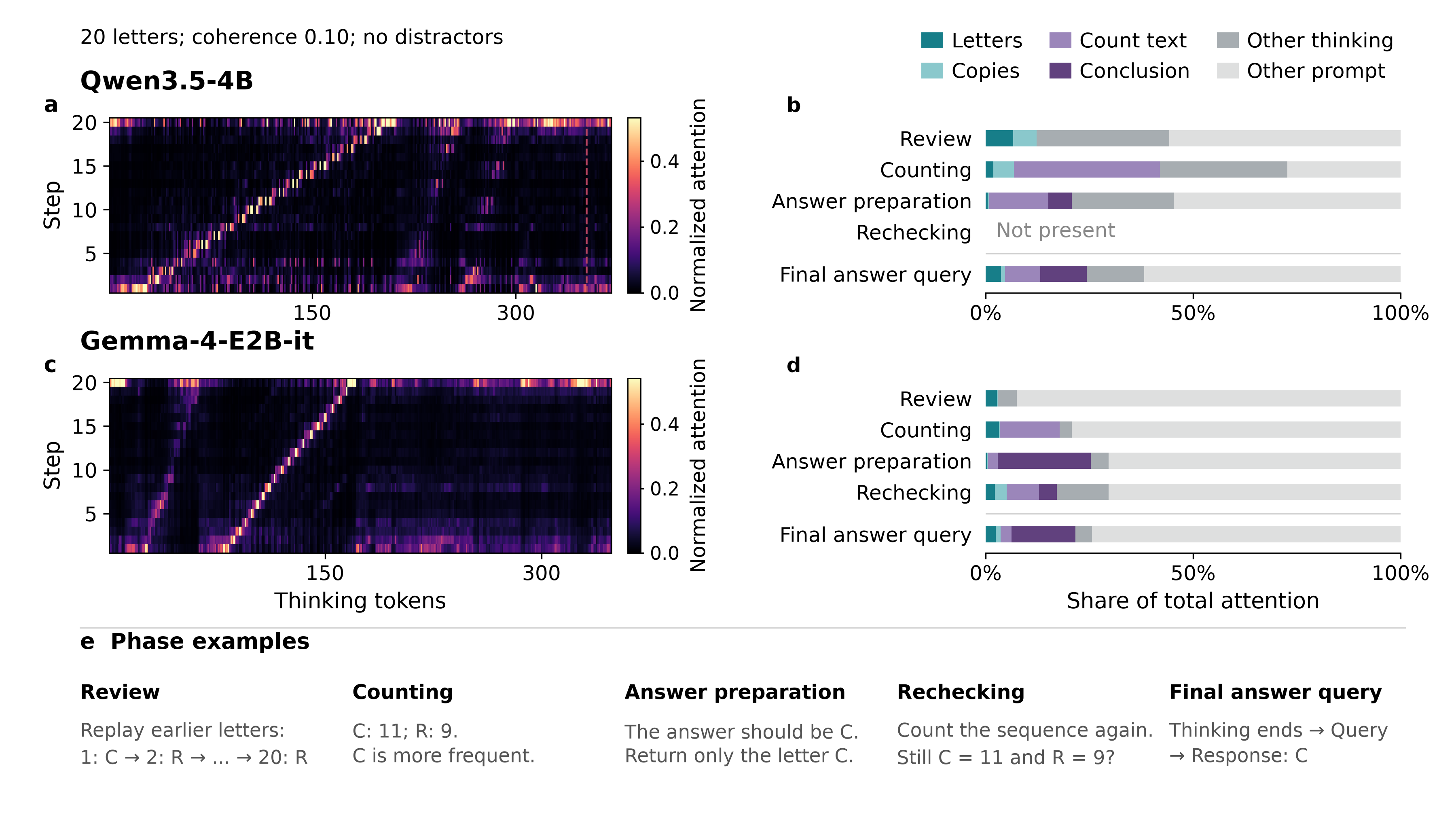}
\caption{\textbf{Selected thinking trajectories revisit the evidence and attend to written counts at the final answer.} (a,c) Attention from each thinking token to the original letters for Qwen3.5-4B and Gemma-4-E2B-it, normalized within each column; dashed line: first expressed preference, the first statement in the generated text that names the eventual answer. (b,d) Trajectories are divided from the generated text into review, counting, answer preparation, and rechecking, with the final answer query summarized separately; bars show absolute shares of total attention to the original letters, letters copied into the reasoning, count text, conclusions, and other context. (e) Condensed example text for each phase. Both trials use 20 letters, coherence 0.10, and no distractors, with the stimulus-selective heads of Figure~\ref{fig:base-performance}e,f (Appendix~\ref{sec:thinking-details}).}
\label{fig:thinking-heatmap}
\label{fig:thinking-phases}
\end{figure}

We selected several 20-letter trials to illustrate how thinking processes the evidence; most other trials we inspected looked similar (Appendix~\ref{sec:thinking-details}). In a 20-letter Gemma-4-E2B-it trial, attention during thinking moved through successive letter positions as the model counted, then swept the sequence again as it rechecked (Figure~\ref{fig:thinking-heatmap}c). A Qwen3.5-4B trial showed the same progression without a rechecking phase (Figure~\ref{fig:thinking-heatmap}a). In both trials, the final answer query directed part of its attention to the written counts and conclusions (Figure~\ref{fig:thinking-phases}b,d). Trials from Qwen3.5-9B and Gemma-4-E4B-it show the same progression (Appendix Figure~\ref{fig:thinking-other-models}).

Together with the failure of direct answers that attend most to the latest letter, these trajectories suggest that thinking constructs explicit intermediate counts for the answer to consult. The kernel and final-query attention can diverge because the kernel reflects the entire computation, whereas final-query attention reflects only its last step. In these trajectories, thinking condenses evidence spread across the sequence into intermediate counts, and the final query attends to those counts. The model thus uses generated text to recreate information already present in its input.

\begin{figure}[!htbp]
\centering
\includegraphics[width=\linewidth]{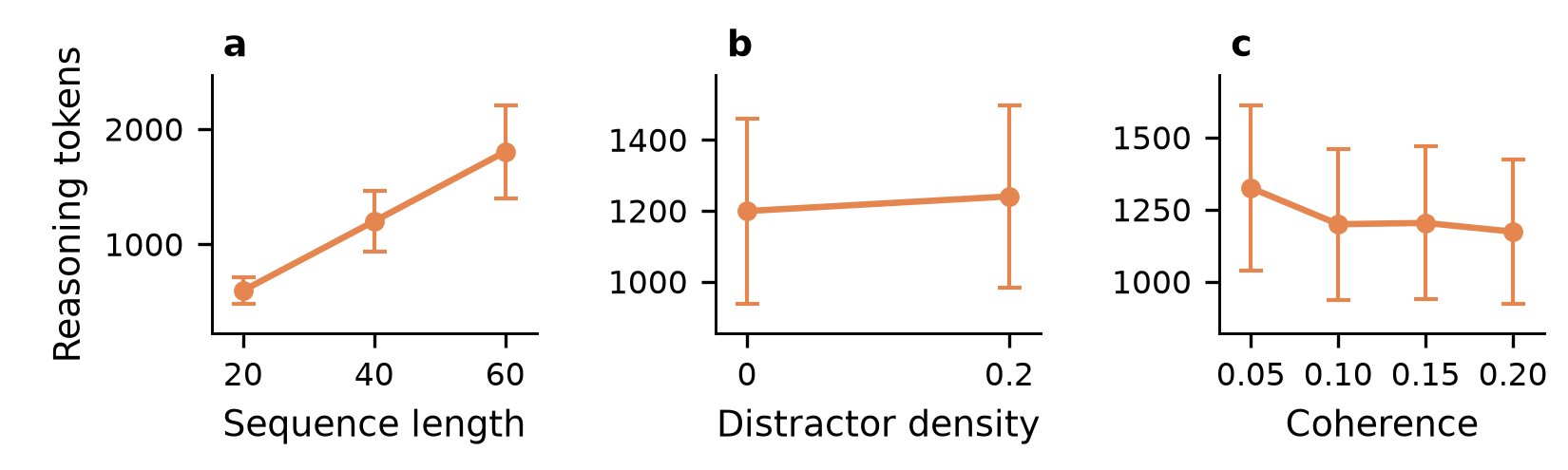}
\caption{\textbf{Reasoning length scales with the number of letters more than with coherence.} Reasoning tokens in thinking trials, excluding answer tokens, by (a) sequence length, (b) distractor density, and (c) coherence over the tested ranges, with the other factors at the baseline. Points show equal-weight means of within-model trial means across seven models; error bars show model SEM ($n=7$ models). Y-axis ranges differ across panels. Trials reaching the token budget are retained.}
\label{fig:reasoning-length}
\end{figure}

This recounting has a computation cost that grows with the number of letters. Reasoning length grew markedly with sequence length in every model (Figure~\ref{fig:reasoning-length}a), shrank more weakly as coherence rose (Figure~\ref{fig:reasoning-length}c), and changed little with distractors (Figure~\ref{fig:reasoning-length}b). The cost thus tracks the number of items to revisit, with a weaker adjustment to the difficulty of the decision. Within the dual-process analogy, this is the price of performing counting in \textit{System~2}: every reasoning token is a sequential generation step, and the cost is paid anew on every trial.

\subsection{In-context reinforcement learning (ICRL) does not teach direct responses to count; they grow more confident but worse}
\label{sec:results-icl}
\begin{figure}[!t]
\centering
\includegraphics[width=\linewidth]{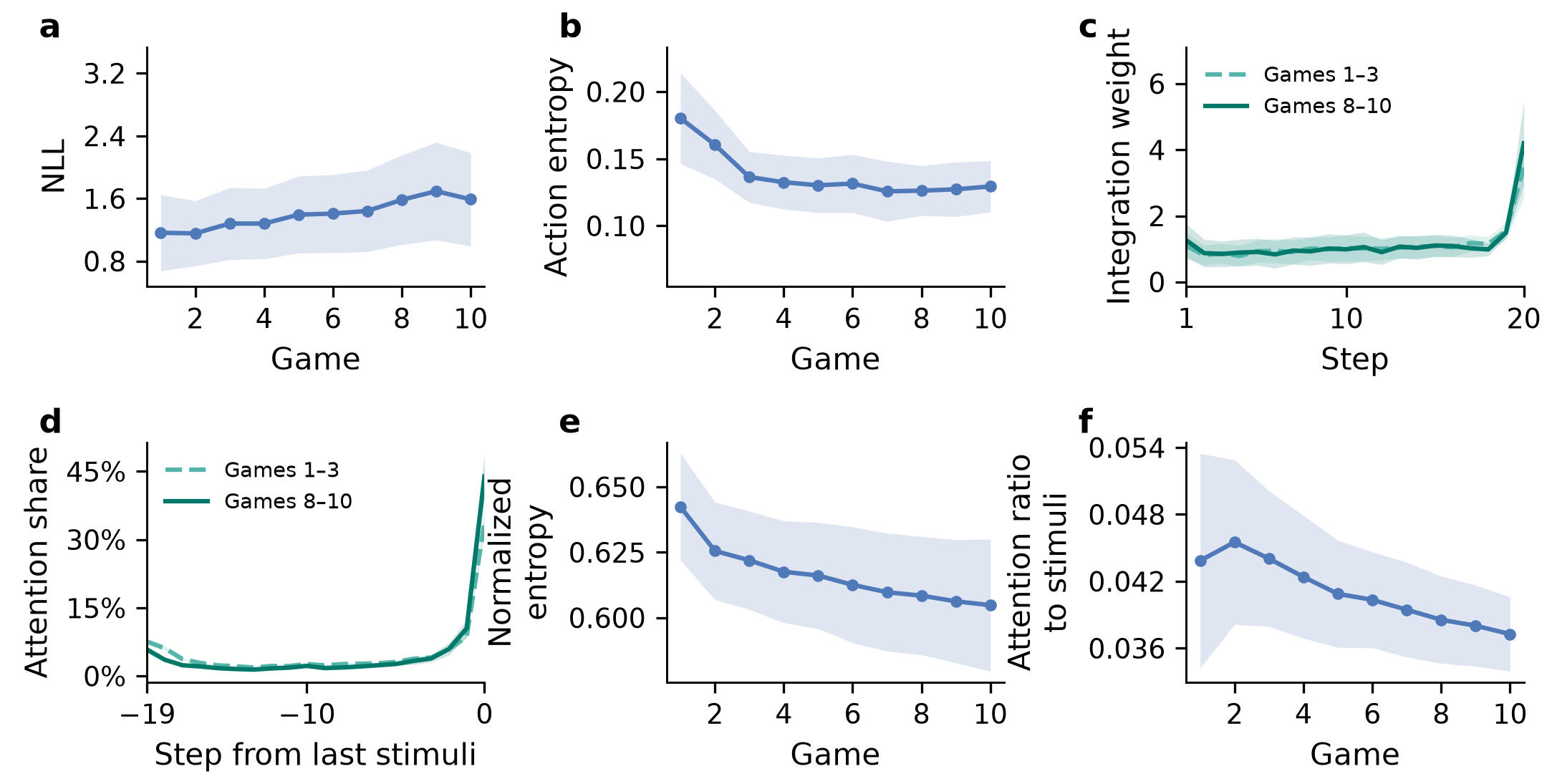}
\caption{\textbf{ICRL does not improve direct evidence integration.} (a) Correct-answer NLL and (b) action entropy by game; (c) integration kernels for initial games (Games 1--3) and final games (Games 8--10). (d) Final-query attention share by lag from the last observed letter (0 = newest; negative values are earlier letters); (e) normalized attention entropy over the full context; (f) fraction of final-query attention on current-game letters. All panels average the two ICRL conditions, with and without distractors. Panels a--c average seven models and d--f the four complete-score models; d,e use stimulus-selective heads and f all heads in the recorded layers. Dashed light teal: initial games; solid dark teal: final games. Bands show model SEM ($n=7$ models in a--c; $n=4$ in d--f). Statistical tests in Appendices~\ref{sec:analysis-details} and~\ref{sec:attention-details}.}
\label{fig:icl-behavior}
\label{fig:icl-attention}
\label{fig:icl-attention-by-game}
\end{figure}

Because LLMs learn task structure statistically in context, including from outcome feedback alone \citep{monea_llms_2024,xiong_large_2026}, repeated games might teach direct responses to integrate the letters; correct-answer NLL would then fall and integration weights would grow more uniform across games. Instead, correct-answer NLL rose across games (Figure~\ref{fig:icl-behavior}a) while action entropy fell (Figure~\ref{fig:icl-behavior}b). As interaction history accumulated, the models thus grew more confident while assigning less probability to the correct answer.

Evidence weighting did not move toward the counting rule as games progressed. In both initial games (Games 1--3) and final games (Games 8--10), kernels showed the recency effect of single-game direct responses of the same length, with weights low across most of the sequence and skewed toward its end (Figure~\ref{fig:icl-behavior}c; Appendix Figure~\ref{fig:base-condition-kernels}a), and the final-letter weight grew in final games (Holm-adjusted $p=0.002$; Appendix~\ref{sec:analysis-details}).

Final-query attention supported the kernel finding of a stronger recency effect in final games: the newest letter received a larger share of stimulus attention (Figure~\ref{fig:icl-attention}d). At the same time, as the conversation grew over games, a smaller fraction of final-query attention reached current-game letters (Figure~\ref{fig:icl-attention}f), with a lower fraction in final than initial games (Appendix Figure~\ref{fig:icl-observed-attention}). Attention entropy over the full context also fell (Figure~\ref{fig:icl-attention}e), indicating more concentrated attention. The final query thus assigned less attention to the current evidence and, within it, more to the newest letter. Across ICRL games, the recency effect strengthened and integration weights remained uneven.

\section{Discussion}
\label{sec:discussion}

\subsection{Simple rules that direct responses do not execute}
Thinking re-executes in text a computation that direct responses do not reliably perform on evidence already in context. Because the counting rule fixes what each letter should contribute to the answer, the kernel shows which letters actually did, and in direct responses the weights skewed toward the end of the sequence. This recency effect parallels those in human free recall \citep{greene_sources_1986}, recency bias in few-shot prompting \citep{zhao_calibrate_2021}, and the position-dependent use of long contexts \citep{liu_lost_2024}. Scratchpads and chain-of-thought improve performance on arithmetic and symbolic problems by making intermediate computation explicit \citep{wei_chainofthought_2022,nye_show_2021}. Our task specifies the contribution of every target occurrence, allowing us to diagnose how thinking changes temporal evidence weighting relative to the exact counting rule. This connects the benefit of intermediate computation to the gap between available information and its effective use, which also arises in in-context learning and working memory \citep{xiong_large_2026,xiong_incontext_2026}. Nor is the gap one of capacity: transformers can express counting \citep{weiss_thinking_2021,yehudai_when_2024}, and interpretability work has found a production model tracking a running character count internally to predict line breaks \citep{gurnee_when_2025}. The relevant boundary thus separates computations that a trained model performs reliably in its direct responses from those it achieves only by unfolding additional steps, and counting, despite its simplicity, falls on the second side.

\subsection{Thinking changes what the final answer reads}
We asked what thinking adds when every letter is already in the context. The evidence favors construction. Thinking made evidence use nearly uniform while the answer's attention to the original letters stayed concentrated on the ends of the sequence. In the selected trajectories, the generated text held counts built by revisiting the input: models constructed counts, compared alternatives, and sometimes checked intermediate conclusions before the final query attended to them. The ordered sweep through the letters resembles forward replay, the reactivation of recent experience in its original order that supports planning in the hippocampus \citep{schuck_sequential_2019,momennejad_offline_2018,mattar_planning_2022}. Unlike neural replay, this sweep leaves a written trace: the generated text serves as a working record that makes accumulated evidence available to subsequent processing, as scratchpads and self-notes do for other algorithmic tasks \citep{nye_show_2021,lanchantin_learning_2023}, and the final query attends to a compact summary of earlier evidence. Two interventions, in the spirit of causal methods in interpretability \citep{geiger_causal_2025,lindsey_biology_2025}, can test this account. Supplying accurate intermediate counts alongside the original sequence, against length-matched uninformative text, would measure how much of the benefit lies in converting distributed evidence into an accessible summary; perturbing the generated counts, or disrupting attention to them, would test whether these intermediate states causally shape NLL and the kernel.

\subsection{Implications for amortization and computation cost}
Resource-rational accounts treat additional computation as worthwhile only when the expected improvement in decision quality outweighs its cost in time and resources \citep{shenhav_expected_2013,griffiths_rational_2015,lieder_resourcerational_2020}. By that standard, a simple and familiar operation such as counting should be performed reliably with little additional computation. Instead, thinking recovered the operation at a computation cost that grew with the number of letters (Figure~\ref{fig:reasoning-length}) and is paid anew on every trial, consistent with the broader observation that scaling thinking can spend substantial computation on simple questions \citep{chen_not_2025}. Amortization describes the alternative: computations that are required frequently can be learned so that their cost is paid largely during learning rather than on every decision \citep{gershman_amortized_2014}. In the brain, replay also serves this function, consolidating offline computation into fast, direct processing \citep{mcclelland_why_1995,barry_neural_2022}; the forward replay performed during thinking leaves the model's weights unchanged, so each trial must repeat the recount. A reliable \textit{System~1} matters for exactly this reason. Performing frequently repeated operations directly saves reasoning tokens and thinking time, leaving extended reasoning for cases where it offers sufficient benefit. If thinking helps by re-encoding evidence into an accessible summary rather than by rereading the input, latent or compressed reasoning should be judged by whether it preserves that summary, and distillation by whether it transfers the construction into the direct pathway rather than only the answers. Experience did not supply such a pathway: across repeated games, ICRL made direct responses more confident but worse, sharpening a policy that still weighted the latest letters most. Systems that accumulate outcome feedback in context may likewise become more decisive without becoming more accurate. Whether an operation that currently requires unfolding can become directly available through learning is the amortization question proper. Answering it requires manipulating learning itself, for example by training or distilling a direct integration policy \citep{yu_distilling_2024,deng_explicit_2024} and comparing, before and after training, direct performance, temporal kernels, the reasoning required to reach a given performance, and generalization to unseen sequences and lengths. Compact working states that update accumulated evidence, retrieval mechanisms that preserve relevant observations, and latent or recurrent computation \citep{geiping_scaling_2025} are candidate implementations, to be evaluated jointly on decision quality, temporal evidence weighting, token use, and measured response time.

\subsection{Scope and limitations}
Letters arrived one per turn and the single-game experiment teacher-forced \texttt{WAIT} at nonfinal steps, so the task probes integration across conversational turns; a single-prompt format could yield a different profile. The mode comparison is empirical. Because Qwen3.5 spans two sampling configurations and cross-model summaries pool complete and top-20 scores, these results describe the tested cohort rather than full-model probabilities. The tested models use at most 12B active parameters per token (Table~\ref{tab:models}), so whether frontier-scale models show the same direct-response gap is untested. Whether the direct-response gap originates in pretraining, post-training, or architecture, for instance in position encodings that index tokens rather than selected content \citep{golovneva_contextual_2024}, and where the boundary between direct and unfolded computation lies for other operations, remains open. A kernel with a recency effect shows a functional temporal bias whose source in input selection, retention, or readout remains open. The intermediate-count account rests on descriptive measurements: attention weights do not by themselves establish causal influence \citep{jain_attention_2019}, the trajectories were selected for illustration and do not estimate how often the process occurs, and the readout account, in which thinking improves access to a count already formed during observation, is not excluded, because we did not test whether the hidden states of direct responses carry a running count across turns; linear probes at each turn and the interventions proposed above would separate these alternatives. Reasoning tokens count sequential generation steps, a measure distinct from FLOPs or latency. In ICRL, NLL and action entropy come from different distributions, so ``more confident'' refers to a more concentrated sampling policy rather than to calibration; task experience, feedback, and context growth also varied together. Because feedback reported only correctness, the experiment tests experience with outcome feedback and leaves open whether the computation performed during thinking can be internalized. Finally, we ran no matched human experiment; the cognitive literature supplies diagnostic methods and candidate processes for this task, with no human benchmark.

\subsection{Conclusion}
Direct responses fail at a computation as simple as deciding which of two letters appeared more often: their integration weights are uneven, skewed toward the end of the sequence and low in its middle, and they do worse as the task becomes harder. Thinking makes evidence weighting nearly uniform while final-query attention to the original letters stays concentrated on the ends of the sequence; in selected trajectories, the model recounts the letters in text and the answer reads those counts, at a cost that grows with the number of letters. ICRL across repeated games makes direct responses more confident but not better. Humans and animals amortize such elementary computations into fast processes, whereas current LLMs pay for them with thinking. These findings pose a clear challenge for future models: determining which elementary computations can be amortized through learning, sparing effortful thinking for problems that need it.

\clearpage

\section*{AI use statement}

We used generative AI tools to polish the manuscript and assist
with writing code for the task and data analysis. We did not
use generative AI tools for any other aspect of this work.
The authors reviewed all AI-assisted work and checked the code
to verify that it performed the intended analyses correctly.
The authors take full responsibility for the final content of
this work, including all AI-assisted text, claims, and artifacts.

\section*{Reproducibility Statement}
We document the experimental procedures in Section~\ref{sec:methods} and Appendix~\ref{sec:suppmethods}, including model checkpoints, experimental conditions, sample sizes, stimulus generation, complete prompts, and generation settings. Appendix~\ref{sec:analysis-details} specifies the behavioral measures, score-coverage and trial-eligibility rules, integration-kernel estimation, and bootstrap procedures, including resampling seeds and multiple-comparison corrections. Appendices~\ref{sec:attention-details} and~\ref{sec:thinking-details} describe attention recording, stimulus-selective head selection, normalization, and the analysis of illustrative thinking trajectories. Appendix~\ref{sec:analysis-reproducibility} documents deterministic seed derivation, trial matching across models and modes, and dataset validation.

\clearpage
\bibliography{references}
\bibliographystyle{iclr2027_conference}

\clearpage
\appendix
\section{Supplementary Methods}
\label{sec:suppmethods}

\subsection{Model checkpoints and experimental conditions}
\label{sec:cohort-conditions}
Tables~\ref{tab:models} and~\ref{tab:conditions} list the seven model checkpoints with their scoring and recorded attention layers, and the complete single-game and ICRL conditions.

For each of the four complete-score models in Table~\ref{tab:models}, the single-game experiment contained 200 stimulus trials per condition, each evaluated in both modes: 2,800 responses per model and 11,200 overall. The 1,400 stimulus sequences were shared across models and modes. ICRL contained 200 ten-game conversation blocks per condition: 4,000 games per model and 16,000 overall. These four models provided complete action scores. The other three models provided returned top-20 token scores, with 4,000 ICRL games each (28,000 games across all seven models). Seven-model single-game and ICRL behavioral summaries combine these scoring schemes; they do not recover full-model probabilities for the top-20 models. Coverage requirements are specified in Appendices~\ref{sec:action-scoring} and~\ref{sec:analysis-details}.

\begin{table}[htbp]
\centering\small
\caption{Model checkpoints, parameter counts, and recorded attention layers. Complete-score models ran locally with Hugging Face Transformers; returned top-20 score models ran through the Tinker API and contributed behavior only. Total parameters include embeddings. Active parameters are those used per token: effective parameters for Gemma-4, which exclude per-layer embeddings, and active-expert parameters for the mixture-of-experts models. Inkling-Small had no public checkpoint and was accessed by name. Layer indices are zero-based transformer-block indices; recorded layers expose explicit full attention.}
\label{tab:models}
\begin{tabularx}{\textwidth}{@{}Xrrl@{}}
\toprule
Checkpoint & Total & Active & Attention layers \\
\midrule
\multicolumn{4}{@{}l}{\textit{Complete-score models (Hugging Face Transformers)}} \\
\texttt{Qwen/Qwen3.5-4B} & 4B & 4B & 3, 7, 15, 23, 31 \\
\texttt{Qwen/Qwen3.5-9B} & 9B & 9B & 3, 7, 15, 23, 31 \\
\texttt{google/gemma-4-E2B-it} & 5.1B & 2.3B & 4, 9, 19, 24, 34 \\
\texttt{google/gemma-4-E4B-it} & 8B & 4.5B & 5, 11, 23, 29, 41 \\
\midrule
\multicolumn{4}{@{}l}{\textit{Returned top-20 score models (Tinker API)}} \\
\texttt{Qwen/Qwen3.6-35B-A3B} & 35B & 3B & --- \\
Inkling-Small & 276B & 12B & --- \\
\texttt{nvidia/\allowbreak NVIDIA-Nemotron-3.5-Lightning-30B-A3B-BF16} & 30B & 3B & --- \\
\bottomrule
\end{tabularx}
\end{table}

\begin{table}[htbp]
\centering\small
\caption{Experimental conditions. $N$: requested horizon; $T$: realized horizon; $D$: distractor count; $g$: target-count difference. Each row had 200 blocks per model. Single-game rows were evaluated in both modes; ICRL rows had ten games per block in non-thinking mode.}
\label{tab:conditions}
\begin{tabular}{@{}lrrrrrr@{}}
\toprule
Experiment & $c$ & $\rho$ & $N$ & $T$ & $D$ & $g$ \\
\midrule
Single-game & .05 & 0 & 40 & 40 & 0 & 2 \\
Single-game & .10 & 0 & 40 & 40 & 0 & 4 \\
Single-game & .15 & 0 & 40 & 40 & 0 & 6 \\
Single-game & .20 & 0 & 40 & 40 & 0 & 8 \\
Single-game & .10 & .20 & 40 & 39 & 8 & 3 \\
Single-game & .10 & 0 & 20 & 20 & 0 & 2 \\
Single-game & .10 & 0 & 60 & 60 & 0 & 6 \\
\midrule
ICRL & .10 & 0 & 20 & 20 & 0 & 2 \\
ICRL & .10 & .20 & 20 & 20 & 4 & 2 \\
\bottomrule
\end{tabular}
\end{table}

\subsection{Stimulus sequence generation}
Each game sampled two distinct target letters uniformly without replacement from the 26 uppercase Latin letters. Sequences were generated by randomly permuting fixed letter counts. Given requested length $N$ and distractor density $\rho$, the distractor count was $D=\operatorname{round}(N\rho)$, rounding to the nearest integer with ties to even, and the requested target count was $L=N-D$. For requested coherence $c$, the target-count difference was $g=\operatorname{round}(cL)$. We set $L^\star=L$ when $L-g$ was even and $L^\star=L-1$ otherwise, giving integer counts
\begin{equation}
n_{\mathrm{dom}}=\frac{L^\star+g}{2},\qquad
n_{\mathrm{other}}=\frac{L^\star-g}{2}.
\end{equation}
Here $n_{\mathrm{dom}}$ is the count of the dominant target letter and $n_{\mathrm{other}}$ that of its competitor; these subscripts describe frequency, not the candidate-order signs used in the integration kernel. The dominant identity was chosen uniformly, and the fixed-count segment was randomly permuted.

Distractor identities were sampled uniformly with replacement from the remaining 24 letters and placed at $D$ uniformly sampled positions in the realized stream. The realized horizon was $T=L^\star+D$. For example, $N=40$, $\rho=0.20$, and $c=0.10$ gave $D=8$, $g=3$, $L^\star=31$, and $T=39$. Requested coherence and density were design parameters; realized proportions could differ after rounding and parity adjustment. Successive stimuli were therefore not independent Bernoulli draws.

The correct answer was the majority target letter across the entire sequence, and every collected sequence had a nonzero count difference.

\subsection{Prompt format and response control}
\label{sec:prompts}
Both experiments used one prompt template, reproduced verbatim below for one trial of each experiment. The system prompt framed the task as a psychophysics experiment, asked the model to determine which target letter occurred more frequently, and told it to disregard all other letters; the prompt calls the target letters ``signal letters''. It also stated the two target-letter proportions, $(1+c)/2$ and $(1-c)/2$, making requested difficulty available to the model. Each stimulus message used the format ``\texttt{Step i/T: X}'', followed by ``\texttt{Your response?}'', where \texttt{i}, \texttt{T}, and \texttt{X} were replaced by the step index, realized horizon, and presented letter; the system prompt stated the same realized horizon. In the single-game experiment, the system prompt named the trial's target letters directly. In ICRL, one system prompt served all ten games of a block and referred to the target letters generically, a user-turn game header announced each game's letters, and a feedback message closed each game.

Thinking and non-thinking evaluations received identical message text. The mode was set only through each model's chat-template thinking switch, and no instruction requested or discouraged reasoning. When a reasoning continuation reached the token budget, only the template's end-of-thinking boundary was inserted before the answer. Every conversation was rendered with the model's own chat template; the boxes below show role-tagged message content without template-specific control tokens. Black assistant turns were inserted into the history, whereas red text marks the model's sampled response or, in feedback, the outcome word (\texttt{correct} or \texttt{incorrect}).

\begin{tcolorbox}[breakable, colback=gray!5!white, colframe=black!75!white, title={Single-game prompt ($N=40$; $c=0.10$; $\rho=0$; target letters E and H)}, fonttitle=\bfseries]
\small
\textbf{<System>}\\
You are a participant in a psychophysics experiment. The experiment runs as a conversation: on each turn you are shown one stimulus (a single uppercase letter), for 40 turns in total. Each user message has two lines: \texttt{`Step \{step\}/40: \{stimulus\}`}, followed by \texttt{`Your response?{}`}.

\smallskip
The signal letters are E and H. Count how many times each appears, then decide which one you received more often. Any other letter is a distractor --- when you see one, simply disregard it.

\smallskip
In this game, one signal letter appears about 0.55 of the time and the other about 0.45. Either letter is equally likely to be the more frequent one. Decide which is more frequent from the stimuli you observe.

\smallskip
After each stimulus, you will be asked \texttt{`Your response?{}`}\\
- If you want to collect more evidence before making a decision, respond WAIT to observe the next stimulus.\\
- If you believe you already know which signal letter is most frequent, respond E or H to give your answer; this ends the current trial early and your answer is final.\\
- On Step 40/40, the trial ends and you must respond with E or H.

\smallskip
Your response must only be E, H, or WAIT.

\vspace{0.8em}
\textbf{<User>}\texttt{Step 1/40: H}\\
\hspace*{3.2em}\texttt{Your response?}\\
\textbf{<Assistant>}\texttt{WAIT}\\
\textbf{<User>}\texttt{Step 2/40: H}\\
\hspace*{3.2em}\texttt{Your response?}\\
\textbf{<Assistant>}\texttt{WAIT}

\vspace{0.3em}
\hspace{1em}$\vdots$
\vspace{0.3em}

\textbf{<User>}\texttt{Step 39/40: E}\\
\hspace*{3.2em}\texttt{Your response?}\\
\textbf{<Assistant>}\texttt{WAIT}\\
\textbf{<User>}\texttt{Step 40/40: E}\\
\hspace*{3.2em}\texttt{Your response?}\\
\textbf{<Assistant>}\textcolor{red}{\texttt{E} or \texttt{H}} \hfill\textit{non-thinking: direct answer}\\
\textbf{<Assistant>}\textcolor{red}{reasoning continuation, then \texttt{E} or \texttt{H}} \hfill\textit{thinking: matched branch}
\end{tcolorbox}

\begin{tcolorbox}[breakable, colback=gray!5!white, colframe=black!75!white, title={ICRL prompt ($N=20$; $c=0.10$; $\rho=0$)}, fonttitle=\bfseries]
\small
\textbf{<System>}\\
You are a participant in a psychophysics experiment. The experiment runs as a conversation containing multiple games. In each game, on each turn you are shown one stimulus (a single uppercase letter), for 20 turns in total. Each stimulus message has two lines: \texttt{`Step \{step\}/20: \{stimulus\}`}, followed by \texttt{`Your response?{}`}.

\smallskip
Before each game, a game header announces its signal letters. Count how many times each appears, then decide which one you received more often. Any other letter is a distractor --- when you see one, simply disregard it.

\smallskip
In each game, one signal letter appears about 0.55 of the time and the other about 0.45. Either letter is equally likely to be the more frequent one. Decide which is more frequent from the stimuli you observe.

\smallskip
After each stimulus, you will be asked \texttt{`Your response?{}`}\\
- If you want to collect more evidence before making a decision, respond WAIT to observe the next stimulus.\\
- If you believe you already know which signal letter is most frequent, respond with one of the current game's signal letters to give your answer; this ends the current trial early and your answer is final.\\
- On Step 20/20, the trial ends and you must respond with one of the current game's signal letters.

\smallskip
Your response must only be one of the current game's signal letters or WAIT.\\
Game headers and feedback:\\
- Reply with \texttt{`OK`} after each game header.\\
- After each game you will be told that the game has ended and whether your answer was correct or incorrect; reply with exactly \texttt{`OK`} and continue to the next game.

\vspace{0.8em}
\textbf{<User>}\texttt{A new game begins. The signal letters you need to count in this game are C and R.}\\
\textbf{<Assistant>}\texttt{OK}\\
\textbf{<User>}\texttt{Step 1/20: C}\\
\hspace*{3.2em}\texttt{Your response?}\\
\textbf{<Assistant>}\textcolor{red}{\texttt{WAIT}, \texttt{C}, or \texttt{R}}

\vspace{0.3em}
\hspace{1em}$\vdots$ \hfill\textit{continues while the model responds \texttt{WAIT}; an earlier letter ends the game}
\vspace{0.3em}

\textbf{<User>}\texttt{Step 20/20: R}\\
\hspace*{3.2em}\texttt{Your response?}\\
\textbf{<Assistant>}\textcolor{red}{\texttt{C} or \texttt{R}}\\
\textbf{<User>}\texttt{This game has ended. Your decision was \textcolor{red}{correct}.}\\
\textbf{<Assistant>}\texttt{OK}\\
\textbf{<User>}\texttt{A new game begins. The signal letters you need to count in this game are J and F.}\\
\textbf{<Assistant>}\texttt{OK}

\vspace{0.3em}
\hspace{1em}$\vdots$ \hfill\textit{Games 2--10 follow the same format}
\end{tcolorbox}

The legal semantic actions were \texttt{WAIT} and the two target-letter choices before the final stimulus, and only the target-letter choices at the final stimulus. In the notation of Appendix~\ref{sec:analysis-details}, the stimulus $\ell_{ti}$ is an observed letter, while an action $a\in\mathcal A_{ti}$ is a response to the observed prefix. The single-game experiment externally inserted \texttt{WAIT} at each nonfinal step; ICRL sampled the legal actions. Inserted acknowledgements and feedback are detailed below.

\subsection{Single-game: paired decisions after complete evidence}
The single-game experiment varied one factor at a time around $c=0.10$, $\rho=0$, and $N=40$ (Table~\ref{tab:conditions}). Every block began with a fresh conversation and contained one game. After each nonfinal stimulus, the assistant response was forced to the canonical text \texttt{WAIT}. The model thus observed the entire sequence; these waits were inserted, not sampled.

At the final stimulus, each trial branched into matched non-thinking and thinking evaluations. Non-thinking selected a legal answer directly. Thinking generated a reasoning continuation at the final decision only, then selected a legal answer. The maximum continuation budget was 4,096 tokens. Branches shared the trial identity, stimulus sequence, and preceding forced-wait history. Mode-specific sampling settings are reported below.

\subsection{ICRL: sequential games with feedback}
Each ICRL block began with a fresh system prompt and contained ten games of one fixed condition. No worked demonstrations preceded the first game. A header announced each game's target letters, followed by an inserted assistant acknowledgement, \texttt{OK}. At each observed stimulus the model sampled an actual response. Waiting advanced the game; a target-letter response ended it immediately and was irrevocable. Waiting was disallowed at the final stimulus, ensuring commitment. Unobserved stimuli after an early commitment were not added to the conversation.

Feedback stated only whether the decision was correct or incorrect. Correctness was evaluated against the preconstructed complete sequence, including unobserved evidence after an early answer. An inserted \texttt{OK} acknowledgement followed feedback. Earlier instructions, headers, observed stimuli, sampled actions, and feedback remained in context for subsequent games. Games within a block were causally dependent, and conversation blocks were the replication units. A change over game index could reflect task experience, the particular feedback history, or growing context; it was not assumed to demonstrate improved performance or isolated rule learning.

\subsection{Action scores and generation settings}
\label{sec:action-scoring}
A model can express the same answer through several tokens, such as \texttt{A}, \texttt{a}, or a space-prefixed \texttt{A}, so each action score pooled all accepted spellings of that action. Fix a trial and a response boundary and suppress both indices: let $o_v$ be the vocabulary logit of token $v$ at that boundary, and $V(a)$ the accepted single-token variants of semantic action $a$. For complete-score models, the action score was
\begin{equation}
s(a)=\log\sum_{v\in V(a)}\exp(o_v).
\end{equation}
Appendix~\ref{sec:analysis-details} restores the indices as $s_{ti}(a)$ for trial $t$ and step $i$. Candidate surfaces were the uppercase and lowercase spellings of each action, bare or preceded by a space or a newline; a surface was accepted only if it encoded to a single token. Each bare uppercase action, including \texttt{WAIT}, was required to be a distinct single token. For a thinking decision, the logits at the final boundary were evaluated after the generated reasoning continuation; the step index $i=T_t$ still refers to the final stimulus, not a reasoning-token index.

Both Qwen3.5 models used temperature $0.7$, top-$k=20$, and top-$p=0.8$ in non-thinking mode; thinking used temperature $1.0$, top-$k=20$, and top-$p=0.95$. Both Gemma models used temperature $1.0$, top-$k=64$, and top-$p=0.95$ in both modes. The presence penalty was $1.5$ for Qwen thinking and $0$ for Gemma; repetition penalties were neutral ($1.0$). Qwen3.6-35B-A3B used the same mode-specific temperature, top-$k$, and top-$p$ as the Qwen3.5 models; Inkling-Small and Nemotron used the default policy of temperature $1.0$, top-$k=50$, and top-$p=0.95$ in both modes. The Tinker API applied no presence or repetition penalty. Penalties applied to multi-token reasoning, not atomic action selection. Thinking-mode sampling settings also governed the final answer.

For Qwen3.6-35B-A3B, Inkling-Small, and Nemotron-3.5-Lightning-30B-A3B-BF16, scores were available only for returned top-20 token surfaces. Accepted returned surfaces were aggregated for each action without recovering omitted mass. Missing scores remained missing. The seven-model single-game and ICRL NLL and kernel panels therefore combine complete-score and truncated-score estimates. They do not establish equivalence between these scoring schemes. All three ran through the Tinker API and contributed behavior only (Table~\ref{tab:models}). The single-game experiment contained 2,800 recorded responses for each model except Nemotron, which had 2,798; kernel fits imposed additional paired-coverage requirements (Appendix~\ref{sec:analysis-details}).

\section{Measurement and Analysis Details}
\label{sec:analysis-details}

\subsection{Behavioral measurements}
We distinguish the presented stimulus from the model's response. For trial $t$, $\ell_{ti}$ denotes the letter presented at step $i$, $T_t$ the realized sequence length, and $a$ a semantic action: wait or choose one of the two target letters. Let $d_t\leq T_t$ be the commitment step and $\mathcal A_{ti}$ the legal action set. Single-game trials always committed at $d_t=T_t$; ICRL could commit earlier. Waiting was legal before the final stimulus and excluded at the final stimulus. We write $s_{ti}(a)$ for an action score and $a_t^\star$ for the correct target-letter action, the action coded by $a^\star$ in Eq.~\ref{eq:reference-count}. Token-to-action aggregation is defined in Appendix~\ref{sec:action-scoring}.

To measure preference for the correct answer, we normalized legal-action scores at unit temperature and computed negative log-likelihood (NLL) at commitment:
\begin{equation}
q_{ti}(a)=\frac{\exp s_{ti}(a)}{\sum_{b\in\mathcal A_{ti}}\exp s_{ti}(b)},\qquad
\mathrm{NLL}_t=-\log q_{t d_t}(a_t^\star).
\end{equation}
For returned top-20 scores, the sum is restricted to observed legal actions; missing correct-answer scores are excluded. This coverage-conditioned normalization does not recover full-model probabilities. Lower NLL indicates greater probability assigned to the correct answer under the corresponding normalization. We separately measured action entropy, $H_t=-\sum_{a\in\mathcal A_{t d_t}}\pi_{t d_t}(a)\log\pi_{t d_t}(a)$, where $\pi_{ti}$ is the probability distribution over legal actions under the executed sampling policy, including temperature and top-$k$/top-$p$ filtering. Lower entropy indicates a more concentrated policy, which need not be more accurate. Both measures included waiting when legal at early commitment. Reasoning length counted generated reasoning tokens, excluding answer tokens and retaining trials that reached the budget. Entropy used natural logarithms.

We report equal-weight model means and model-level standard errors unless specified otherwise. The exploratory final-minus-initial game comparisons (Games 8--10 minus Games 1--3) in ICRL resampled whole conversation blocks, holding the model cohort fixed; the grouped integration kernels used a whole-curve bootstrap comparison. The paired procedures are specified below.

\subsection{Temporal integration kernels}
\label{sec:kernel-fitting}
The kernel asks how far a single target letter at each position moves the final preference toward that letter, holding the letters at other positions fixed. Each fit used one cell, defined by a model, condition, and single-game mode or by a model, condition, and ICRL game index; fix such a cell and suppress these indices. Let $t=1,\ldots,n$ index its eligible trials and $i=1,\ldots,T$ its stimulus positions, where $T$ is the realized sequence length. We coded $x_{ti}=-1$ for the first target letter in the recorded candidate order, $+1$ for the second, and $0$ for a distractor. Let $a_t^{(1)}$ and $a_t^{(2)}$ denote the corresponding choice actions. The response variable was their final score difference, $y_t=s_{tT}(a_t^{(2)})-s_{tT}(a_t^{(1)})$. Candidate order was independent of correctness, and because $x_{ti}$ and $y_t$ share one orientation, treating the second candidate as A in Eq.~\ref{eq:reference-count} leaves $w_i$ unchanged.

We estimated the intercept $b$ and position weights $\mathbf w=(w_1,\ldots,w_T)$ by minimizing squared prediction error with an L2 penalty on the weights:
\begin{equation}
(\widehat b,\widehat{\mathbf w})=
\operatorname*{arg\,min}_{b,\mathbf w}
\left\{
\underbrace{\sum_{t=1}^{n}\left(y_t-b-\sum_{i=1}^{T}w_i x_{ti}\right)^2}_{\text{squared prediction error}}
+\underbrace{\lambda\sum_{i=1}^{T}w_i^2}_{\text{weight penalty}}
\right\}.
\label{eq:integration-kernel}
\end{equation}
The optimized variables were $b$ and $\mathbf w$; the regularization strength $\lambda$ was fixed at 1, not fitted. The intercept was not penalized. The estimated coefficients $\widehat w_i$ formed the temporal integration kernel. A positive coefficient means that a target at position $i$ was associated with a final score difference favoring that target.

Single-game mode comparisons used identical eligible trial IDs within each model and condition, requiring complete observation and finite scores for both target actions in both modes. These paired subsets were not intersected across models, and missing scores were not imputed. ICRL kernels were fit separately by game using only games that observed all 20 letters; about $12\%$ ended earlier and were excluded. We compared initial games (Games 1--3) with final games (Games 8--10), weighting games equally within each stage.

Single-game integration kernels were displayed by requested horizon (20, 40, or 60 letters) and retained native steps, including the realized 39-step distractor condition. The seven-model condition-specific kernels used mode-paired coverage subsets within each model and condition. Across the seven conditions, the paired sample size was 140--167 for Nemotron, 195--200 for Qwen3.6, and 200 for each of the other five models; in the 40-letter baseline of Figure~\ref{fig:base-performance}d, they were 156 for Nemotron and 195 for Qwen3.6. Within each model, conditions were averaged equally; game-stage curves then averaged initial games or final games with equal game weights. No score-scale normalization was applied across models.

Accumulator models of perceptual decisions offer one interpretation of a behavioral kernel \citep{usher_time_2001,brunton_rats_2013}. A simple leaky-accumulator description expresses the mean dynamics of a decision variable $z(t)$ as
\begin{equation}
\frac{\mathrm{d}z(t)}{\mathrm{d}t}=-\frac{z(t)}{\tau}+w(t)e(t),
\label{eq:intro-accumulator}
\end{equation}
where $e(t)$ is signed evidence, $w(t)$ weights incoming evidence, and $\tau$ sets how long accumulated evidence is retained. The weight $w(t)$ is the continuous counterpart of the integration weights $w_i$ in Eq.~\ref{eq:integration-kernel}. With zero initial state, an observation at time $t$ contributes to the final state at $T$ in proportion to $w(t)\exp[-(T-t)/\tau]$, so the kernel recovers $w(t)$ directly only when $\tau\to\infty$. The task determines the optimal values of both factors. Counting across a whole sequence requires constant $w(t)$ and $\tau\to\infty$, the continuous counterpart of Eq.~\ref{eq:reference-count}; if only one interval matters, $w(t)$ should select that interval; and in changing environments, a finite $\tau$ discounts evidence that has become outdated \citep{glaze_normative_2015}. A terminal kernel, however, measures only this product of input weighting and retention and cannot identify either factor separately, and uneven temporal weighting can also arise from other mechanisms \citep{keung_divisive_2020}. We therefore interpret the kernels in the main text as functional descriptions of temporal weighting rather than as estimates of leak.

\subsection{Paired comparisons and uncertainty}
For the grouped kernels in Figure~\ref{fig:icl-behavior}c, we tested the final-minus-initial game difference (Games 8--10 minus Games 1--3) across all 20 positions. This comparison matched the seven-model behavioral panel, averaging its two conditions equally within model and weighting models and games equally. Four models supplied complete action scores; three supplied returned top-20 scores, with missing final target-action pairs excluded rather than imputed. We drew 9,999 conversation-block bootstrap samples (seed 20260925, incremented by condition index), independently within each condition and jointly across games and stimulus-matched models. In each resample we refit the same ridge estimator after applying each game's full-horizon and score-coverage eligibility. Let $\widehat{\boldsymbol\delta}$ be the 20-position difference between the two grouped mean kernels. The whole-curve statistic was $Q=\|\widehat{\boldsymbol\delta}\|_2^2$; the centered bootstrap null statistics were $Q^*=\|\boldsymbol\delta^*-\widehat{\boldsymbol\delta}\|_2^2$. We computed $p=(1+\#\{Q^*\geq Q\})/10{,}000$ for this single exploratory comparison, without treating positions as independent replicates. The whole-curve test gave $p=0.0001$ (no centered-null exceedance), and the four-complete-score sensitivity analysis gave the same value. Among position-wise follow-up tests, only the final position differed after Holm correction, with a larger weight in Games 8--10 (Holm-adjusted $p=0.002$). Position-wise follow-up tests used two-sided centered bootstrap errors, $p_i=(1+\#\{|\delta_i^*-\widehat\delta_i|\geq|\widehat\delta_i|\})/10{,}000$, with Holm correction across the 20 positions. A sensitivity analysis used the four complete-score models with the same draws and a separate 20-position Holm family. Saved percentile confidence intervals are pointwise, not multiplicity-adjusted. The test retains the raw score scales and can be dominated by large weights on the final letters. Inference is conditional on the fixed model cohort and game-specific eligible samples; it is not an equivalence test or a causal test of history length. Figure bands remain model SEM ($n=7$ models), not bootstrap confidence intervals.

\subsection{Attention measurement and head selection}
\label{sec:attention-details}
Explicit attention weights were recorded at the assistant boundary immediately before an action. For the final thinking decision, this query followed the reasoning continuation. Five eligible full-attention layers were recorded per model (Table~\ref{tab:models}); when the inference backend did not return attention weights, the same causal prefix was passed through the model again with explicit (eager) attention to record them. Each head was normalized over all visible keys, including the query token. Population measurements cover these recorded attention layers. The same selected heads are applied to the single-trial thinking examples in Appendix~\ref{sec:thinking-details}.

Stimulus attention was the summed mass on original stimulus-letter tokens. In ICRL this target included only observed letters from the current game, while the denominator included the full conversation. Final-query comparisons used the actual answer boundary, including early commitments.

Because individual heads specialize, averaging over all heads would mix attention directed at the letters with attention serving other functions. Every single-game attention panel therefore uses stimulus-selective heads, the small set carrying most attention to the letters, chosen on held-out conversation blocks so that selection and evaluation never share a trial. Figure~\ref{fig:base-performance}e,f averages raw attention over selected heads within each trial, normalizes over original stimuli, then averages trials and the four models equally; coherence is 0.10, with no distractors and 40 letters. Blocks were divided into two folds by block-index parity. Within each training fold, attention to original stimulus letters in thinking trials was averaged over trials within condition, then over conditions equally. Heads were ranked across the five recorded layers, and the smallest set accounting for at least 90\% of training target mass was retained. Held-out thinking and non-thinking trials used the same opposite-fold roster. Selected heads were pooled across layers before trial and condition averaging.

ICRL heads were selected independently using the same 90\% cumulative-mass rule. Within each opposite training fold, final-query attention to original current-game stimuli was averaged over trials within each condition/game and then equally across condition/game cells, including those of two collected ICRL conditions not analyzed here. A single global roster across the five recorded layers was retained for every game of each held-out conversation block; heads were not reselected for initial versus final games. This selected 15 heads per Qwen model and 17 per Gemma model in each fold. The 90\% threshold refers to training attention mass, not the proportion of heads. As an all-head comparison, Figure~\ref{fig:icl-attention}f pools all heads across the five recorded layers at the final answer query, averages trials within each condition/game and then the two conditions equally within each model, and reports the equal-weight model mean and model-level SEM ($n=4$ models).

Scalar summaries pooled selected heads across layers within each trial, rather than giving layers with different selected-head counts equal weight. Normalized attention entropy measures how widely a head spreads its attention over the visible context, from 0 when all attention falls on one key to 1 when attention is uniform. For each selected head with key probabilities $p_j$, it was
\begin{equation}
H_{\mathrm{norm}}=\frac{-\sum_{j=1}^{K}p_j\log p_j}{\log K},
\end{equation}
where $K$ was the number of visible keys. Entropy was calculated before head averaging.

ICRL within-game curves measured attention to current-game stimulus letters at each observed response boundary; missing post-commitment positions were excluded. Final-query stimulus-position profiles first averaged raw attention equally across selected heads within each trial, then normalized over observed current-game stimulus letters, matching the single-game profile estimator. Profiles were aligned by lag from the last observed stimulus and displayed oldest to newest across 20 columns. Its position labels correspond to absolute positions only for full-horizon games; unavailable earlier lags were missing rather than zero. Each trial profile summed to one, but the plotted pointwise means need not sum to one because available trials differ across lags. Game-stage summaries averaged initial games (Games 1--3) and final games (Games 8--10).

For Figure~\ref{fig:icl-attention}e,f, exploratory paired comparisons tested final games minus initial games (Games 8--10 minus Games 1--3) in the fixed four-model attention cohort. Each test matched its plotted estimator: normalized full-context entropy averaged over frozen opposite-fold mass90 heads (e), and final-query current-game stimulus mass averaged over all heads in the five recorded layers (f), each in the two ICRL conditions. All 200 conversation blocks per condition contributed all ten games. We averaged games within each group, conditions within model, and models equally. We used 49,999 whole-conversation bootstrap resamples (seed 20260918), independently within condition with shared draws across games and stimulus-matched models. We report percentile 95\% confidence intervals and two-sided centered bootstrap $p$-values, computed as $(1+\#\{|\Delta^*-\widehat\Delta|\geq|\widehat\Delta|\})/50{,}000$; Bonferroni correction covered these two pooled attention tests. Normalized attention entropy decreased from 0.630 in Games 1--3 to 0.607 in Games 8--10 (paired change: $-0.02346$; 95\% CI: $[-0.02406,-0.02286]$), and the current-game stimulus fraction decreased from 0.0445 to 0.0379 (paired change: $-0.00652$; 95\% CI: $[-0.00686,-0.00618]$); both gave Bonferroni-adjusted $p=0.00004$. Head rosters remained fixed in every resample, so uncertainty is conditional on the selected heads and model cohort. These grouped tests do not test monotonic trends or causal effects; figure bands remain model SEM ($n=4$ models).

\subsection{Thinking heatmaps and phase annotations}
\label{sec:thinking-details}
Thinking-token heatmaps and phase-allocation panels described selected single-game trials from the four primary models. Their attention was recorded by passing each saved trial, including its generated reasoning, through the model again. These recordings covered eight explicit attention layers for Qwen and seven for Gemma, but analysis retained only the frozen single-game roster from the opposite block fold, mapped by layer/head identity into the recorded layers. No heads were selected using the displayed trajectory. For each consumed thinking token, attention to each original step was averaged equally over these selected heads, then normalized over original steps. Thus each heatmap column shows the distribution conditional on attending to original stimuli, with uniform share $1/T$. The heatmaps excluded prompt-final and post-thinking answer queries. Color limits used the within-trial 99.5th percentile and could differ across examples.

Section boundaries were marked from the generated text. The preference marker identified the first detected verbal statement naming the recorded thinking choice; it was not an inferred internal commitment time. Reviewed text anchors divided each example into phases such as reviewing the request and stimuli, counting and comparing, and preparing or rechecking the answer. The final answer query was summarized separately. We selected the displayed trials (Figures~\ref{fig:thinking-heatmap} and~\ref{fig:thinking-other-models}) for illustration; most other trials we inspected showed the same qualitative progression, with the model revisiting the letters and writing out intermediate counts before answering.

For phase-allocation bars, visible keys were partitioned into original stimulus letters, verified copied letter tokens, counting text, conclusion text, other thinking text, and other prompt tokens. Copied letters were checked against the original sequence and numbered positions; counting and conclusion categories used reviewed whole-line or exact-span annotations. Category masses were summed within each normalized head, then averaged equally across the selected heads and query tokens in that phase. Unlike the heatmaps, these bars retained absolute shares of total attention. Phase names and annotations were descriptive and specific to each example; no population uncertainty was assigned to these single-trial displays.

\subsection{Dataset validation and reproducibility}
\label{sec:analysis-reproducibility}
A study seed of 42 initialized deterministic, domain-separated stimulus, action, and thinking seed derivations. Trial identity incorporated condition, block, and game, allowing stimulus matching across models and paired single-game modes. ICRL games remained within their original conversations. Analyses used validated manifests and trial metadata from the designated single-game and ICRL datasets; partial blocks and historical cohorts were excluded.

\clearpage
\section{Supplementary Results}
\label{sec:supp-results}
These figures extend the seven-model behavioral and four-model attention results in Section~\ref{sec:results}. Error bars and bands denote one model-level SEM ($n=7$ models for kernels; $n=4$ for attention) unless the caption identifies a descriptive display. Narrow SEM bands may be obscured by the mean curves at the plotted scale.

\subsection{Single-game integration kernels by model}
\begin{figure}[!htbp]
\centering
\begin{minipage}[t]{.325\linewidth}
\panelgraphic{a}{\includegraphics[width=\linewidth]{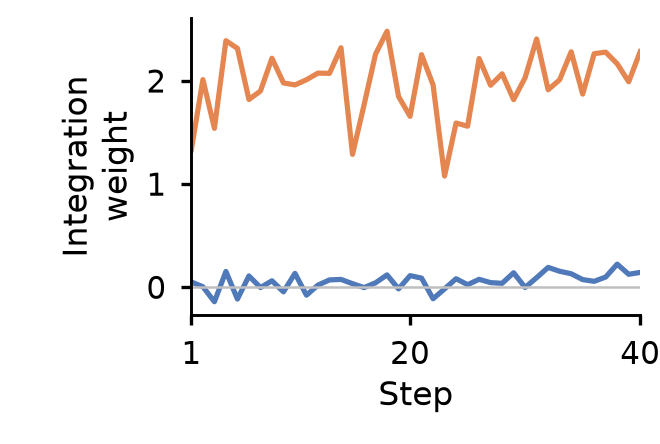}}
\end{minipage}\hfill
\begin{minipage}[t]{.325\linewidth}
\panelgraphic{b}{\includegraphics[width=\linewidth]{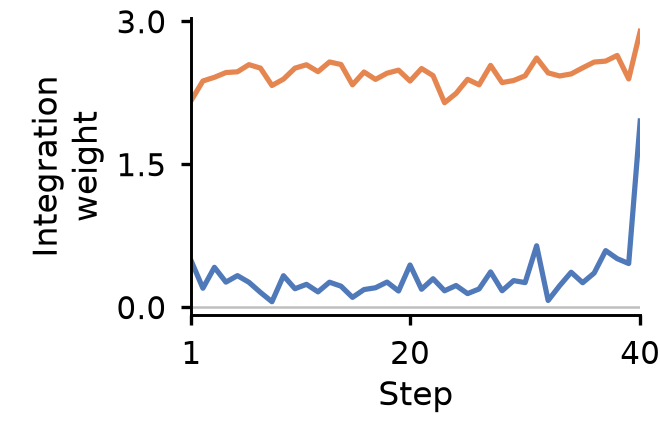}}
\end{minipage}\hfill
\begin{minipage}[t]{.325\linewidth}
\panelgraphic{c}{\includegraphics[width=\linewidth]{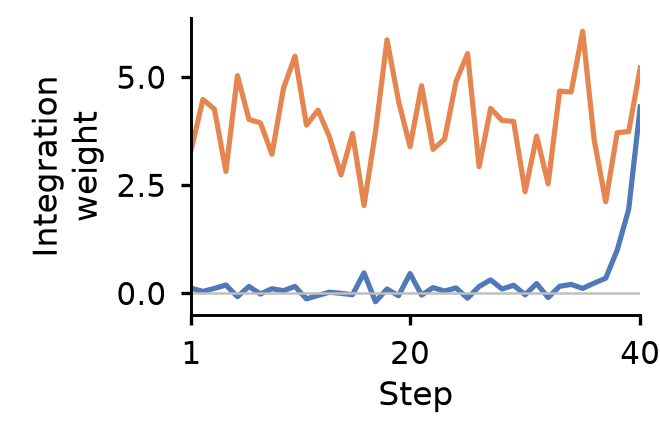}}
\end{minipage}\par\medskip
\begin{minipage}[t]{.325\linewidth}
\panelgraphic{d}{\includegraphics[width=\linewidth]{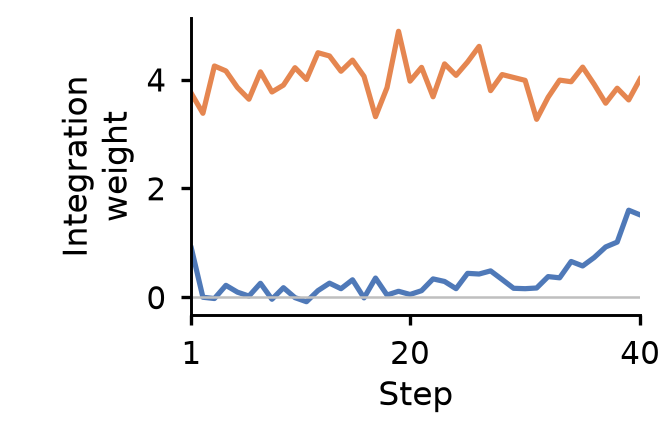}}
\end{minipage}\hfill
\begin{minipage}[t]{.325\linewidth}
\panelgraphic{e}{\includegraphics[width=\linewidth]{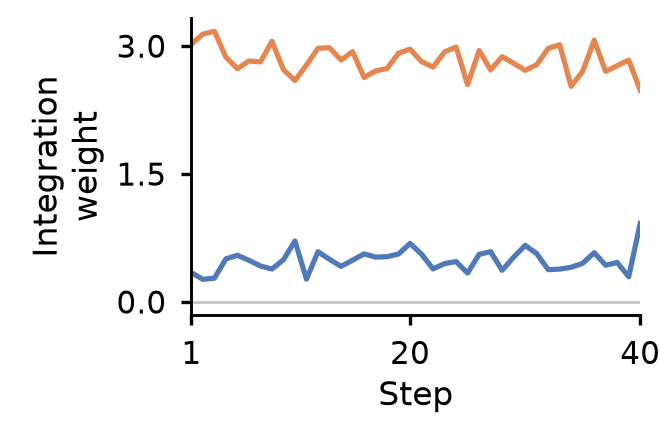}}
\end{minipage}\hfill
\begin{minipage}[t]{.325\linewidth}
\panelgraphic{f}{\includegraphics[width=\linewidth]{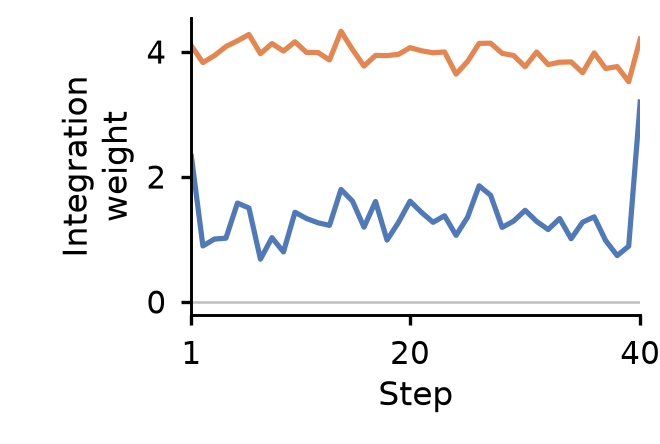}}
\end{minipage}\par\medskip
\begin{minipage}[t]{.325\linewidth}
\panelgraphic{g}{\includegraphics[width=\linewidth]{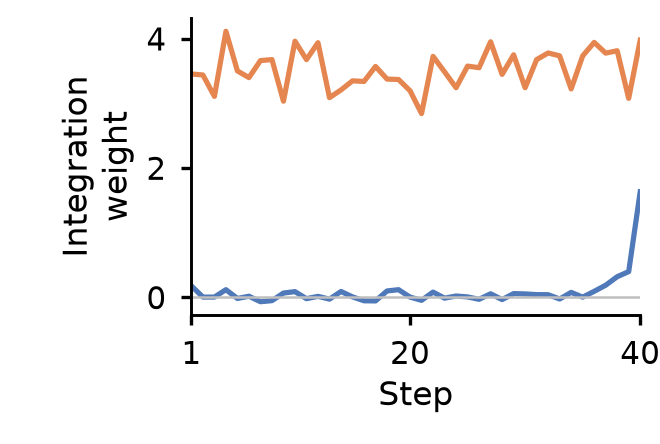}}
\end{minipage}\par\medskip
\caption{\textbf{Individual-model integration kernels for the single-game mode comparison.} Kernels underlying Figure~\ref{fig:base-performance}d, with 40 letters, coherence 0.10, and no distractors: (a) Qwen3.5-4B, (b) Qwen3.5-9B, (c) Gemma-4-E2B-it, (d) Gemma-4-E4B-it, (e) Qwen3.6-35B-A3B, (f) Inkling-Small, and (g) Nemotron-3.5-Lightning-30B-A3B-BF16. Blue: non-thinking; orange: thinking. Both modes use the same eligible trials within each model, with 200 paired trials per model in a--d and f, 195 in e, and 156 in g; eligibility is not intersected across models. Ridge fits ($\lambda=1$) retain raw target-action score differences without cross-model normalization, and vertical scales vary across panels. For e--g, both target-letter scores must be present among the returned top-20 scores. Curves are point estimates without within-model uncertainty bands.}
\label{fig:individual-base-kernel}
\end{figure}

The seven individual-model kernels in Figure~\ref{fig:individual-base-kernel} disaggregate the baseline mode comparison in Figure~\ref{fig:base-performance}d, retaining each model's raw score scale and paired eligible trials.

\clearpage

\subsection{Temporal integration kernels across single-game conditions}
\begin{figure}[htbp]
\centering
\begin{minipage}[t]{.49\linewidth}
\panelgraphic{a}{\includegraphics[width=\linewidth]{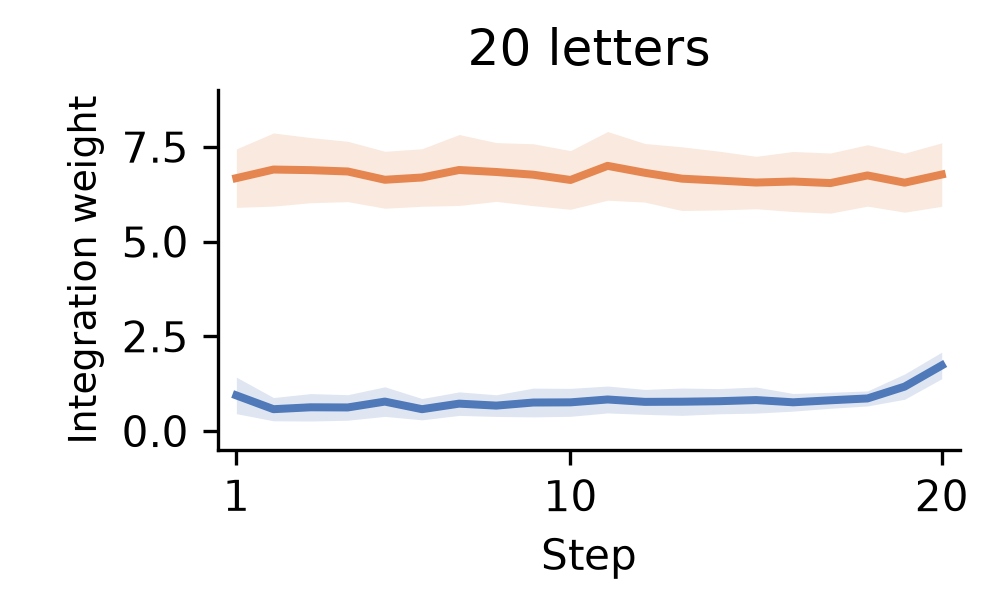}}
\end{minipage}\hfill
\begin{minipage}[t]{.49\linewidth}
\panelgraphic{b}{\includegraphics[width=\linewidth]{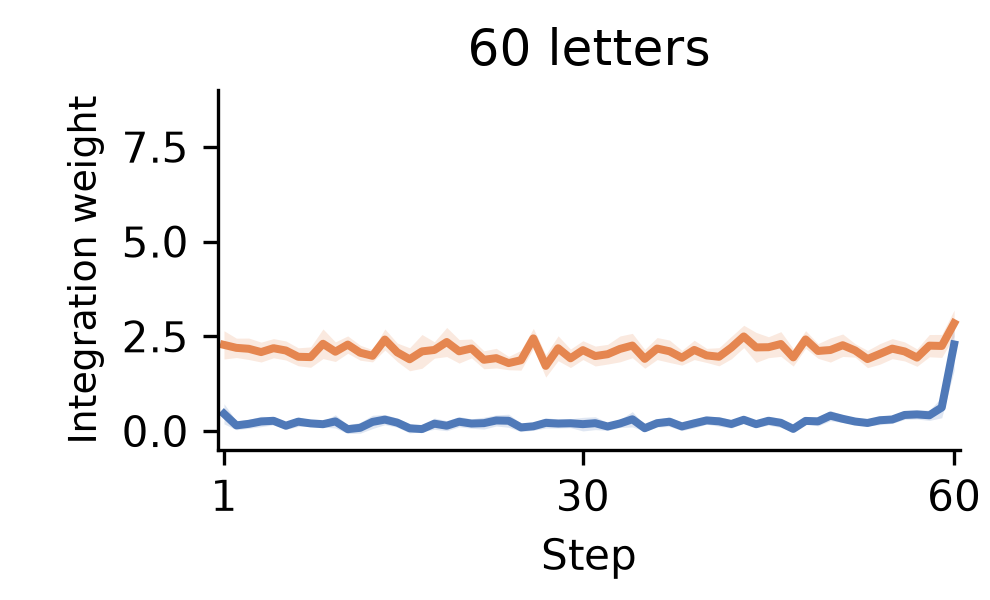}}
\end{minipage}\par\medskip
\begin{minipage}[t]{.49\linewidth}
\panelgraphic{c}{\includegraphics[width=\linewidth]{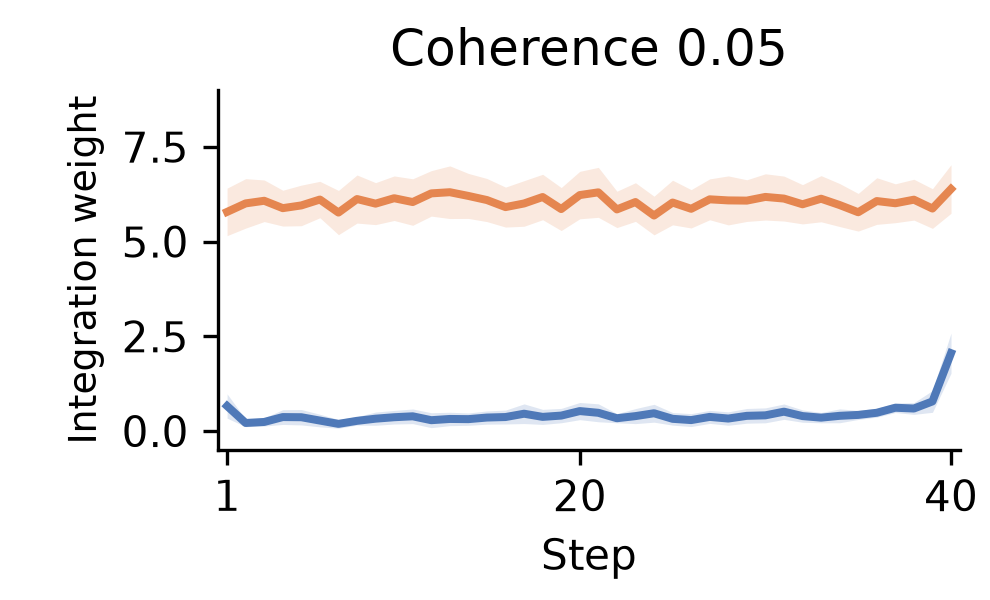}}
\end{minipage}\hfill
\begin{minipage}[t]{.49\linewidth}
\panelgraphic{d}{\includegraphics[width=\linewidth]{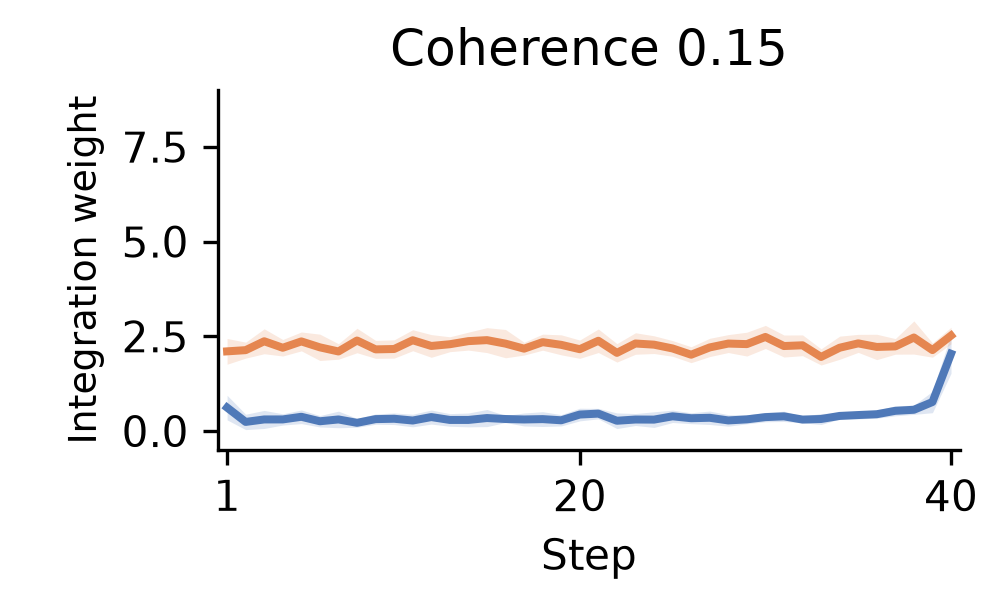}}
\end{minipage}\par\medskip
\begin{minipage}[t]{.49\linewidth}
\panelgraphic{e}{\includegraphics[width=\linewidth]{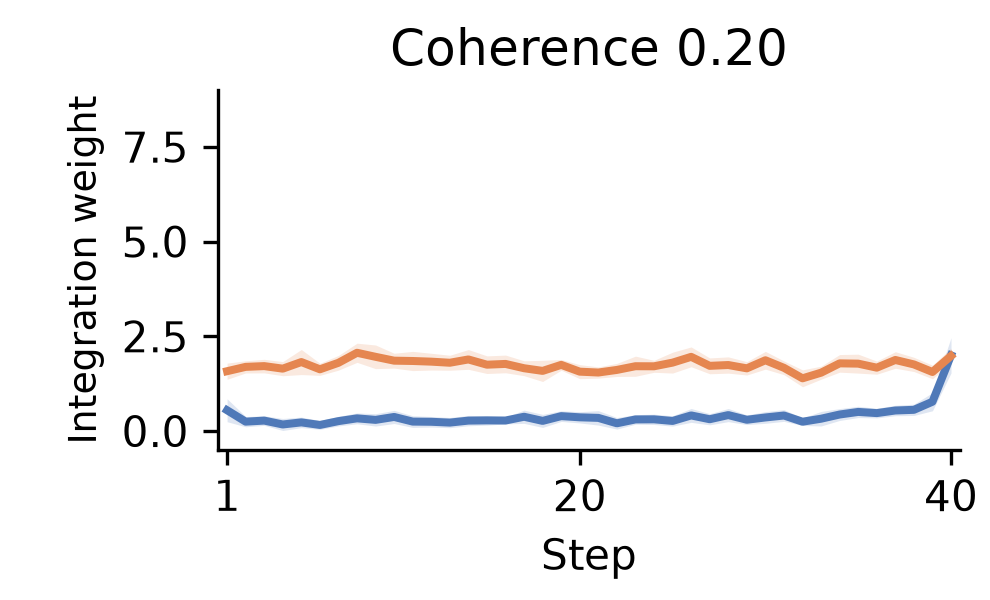}}
\end{minipage}\hfill
\begin{minipage}[t]{.49\linewidth}
\panelgraphic{f}{\includegraphics[width=\linewidth]{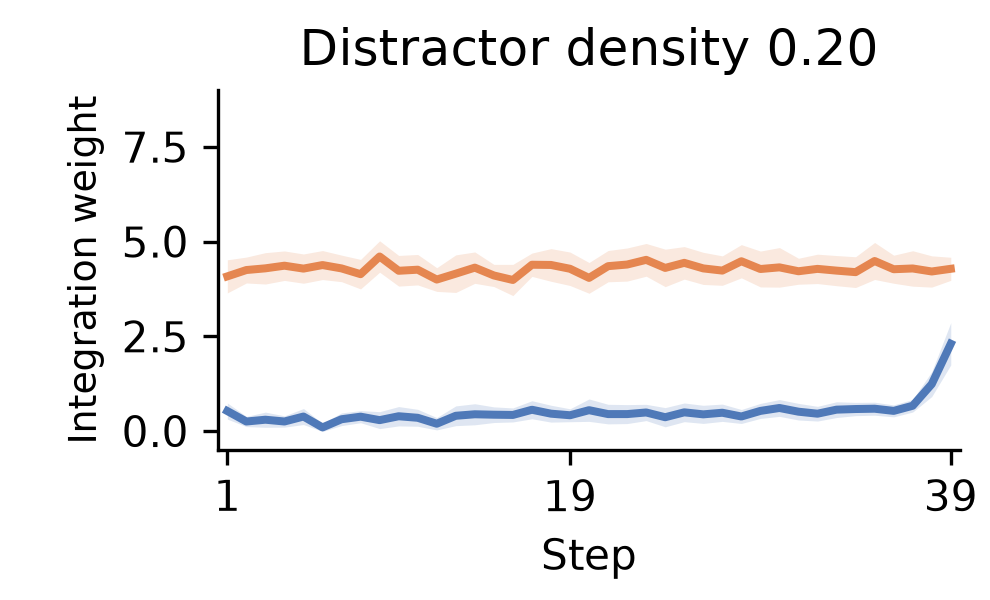}}
\end{minipage}\par\medskip
\caption{\textbf{Non-thinking kernels keep their recency effect in every single-game condition.} Seven-model temporal integration kernels for the remaining single-game conditions. (a,b) Sequence lengths 20 and 60; (c--e) coherence 0.05, 0.15, and 0.20; (f) distractor density 0.20. Other factors are fixed at 40 letters, coherence 0.10, and no distractors. Panel f retains the realized 39 letter positions. Blue: non-thinking; orange: thinking. Curves show equal-weight model means with model SEM bands ($n=7$ models). Both modes use the same eligible trials within each model and condition, without a cross-model trial intersection. For the three returned top-20 score models, fits use the returned target-letter scores. All panels use raw action-score differences and a shared vertical scale, which is wider than that of Figure~\ref{fig:base-performance}d.}
\label{fig:base-condition-kernels}
\end{figure}

Across the other six single-game conditions, non-thinking kernels keep their recency effect and thinking kernels stay nearly uniform (Figure~\ref{fig:base-condition-kernels}), extending the baseline comparison in Figure~\ref{fig:base-performance}d.

\clearpage

\subsection{Final-query stimulus attention across sequence lengths}
\begin{figure}[htbp]
\centering
\panelgraphic{a}{\includegraphics[width=\linewidth]{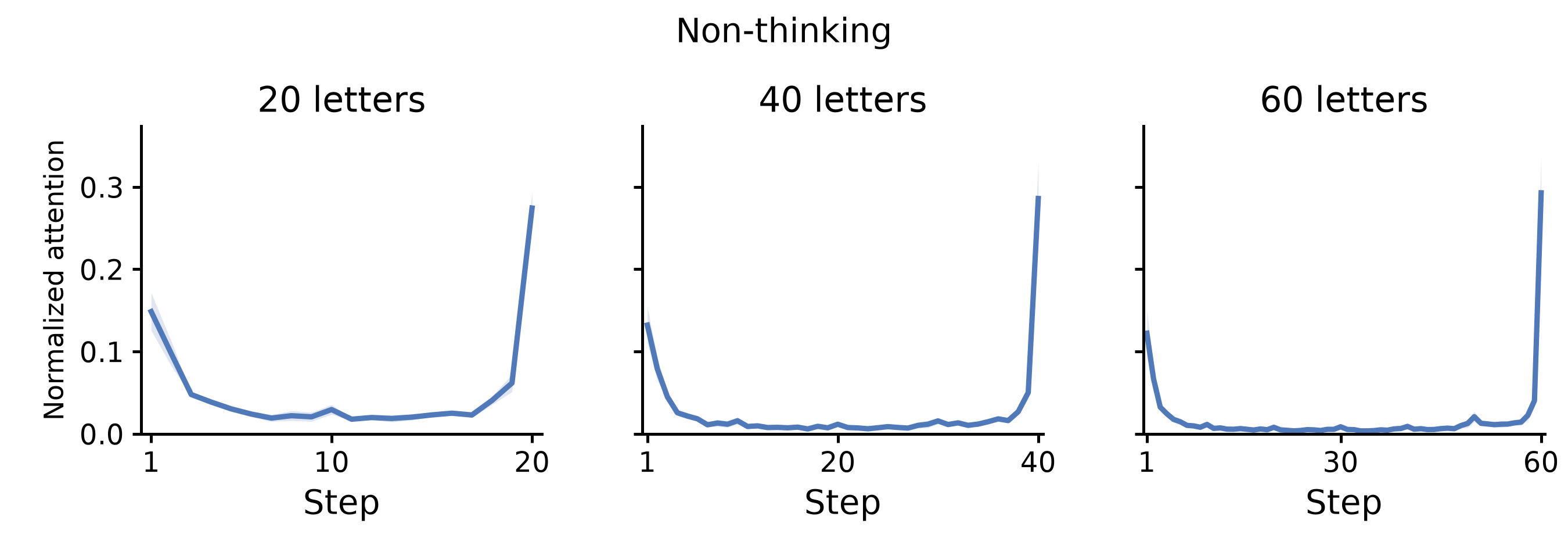}}
\par\medskip
\panelgraphic{b}{\includegraphics[width=\linewidth]{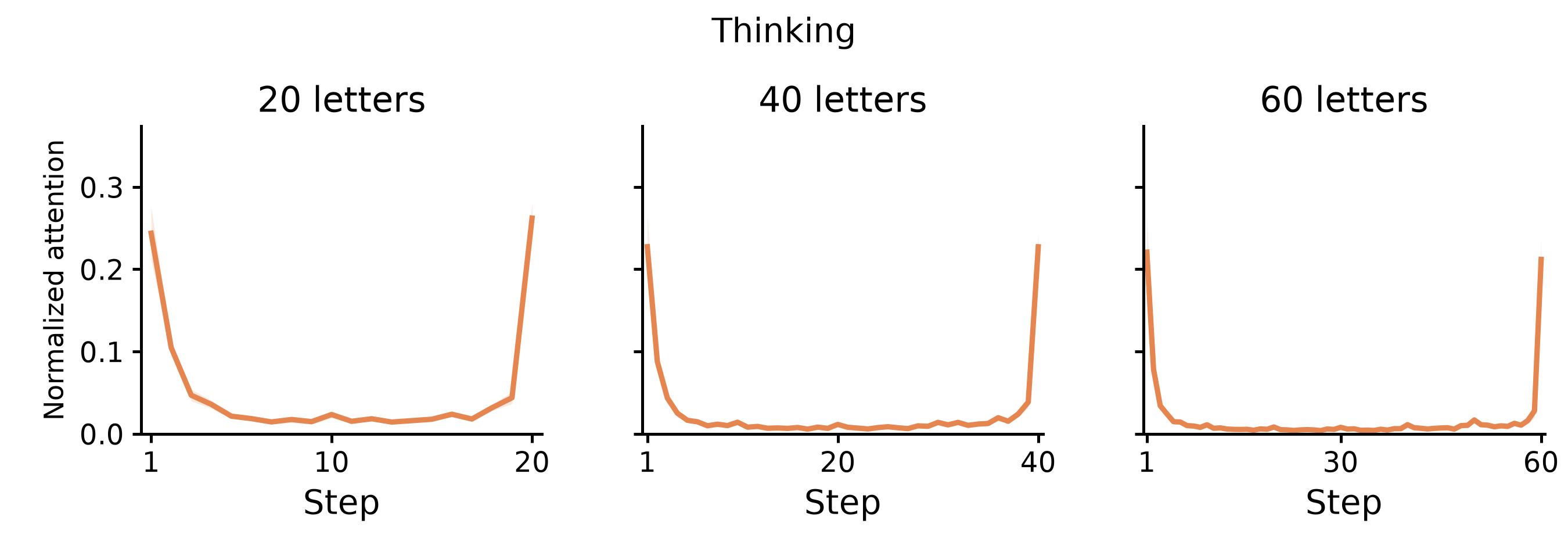}}
\caption{\textbf{Final-query attention stays concentrated on the sequence ends at every tested length.} Final-query stimulus attention across sequence lengths in four models. (a) Non-thinking; (b) thinking. Columns show 20, 40, and 60 letters, all at coherence 0.10 without distractors. Both modes use the same frozen, thinking-trained opposite-fold mass90 heads. Within each trial, raw selected-head attention is averaged before normalization over original stimulus letters, then averaged across trials and equally across models. Each curve sums to one; bands show model SEM ($n=4$ models). The 40-letter profiles reproduce Figure~\ref{fig:base-performance}e,f.}
\label{fig:base-attention-lengths}
\end{figure}

At 20, 40, and 60 letters alike, final-query attention concentrates on the first and last letters in both modes (Figure~\ref{fig:base-attention-lengths}).

\clearpage

\subsection{ICRL final-query attention across game stages}
\begin{figure}[htbp]
\centering
\includegraphics[width=.42\linewidth]{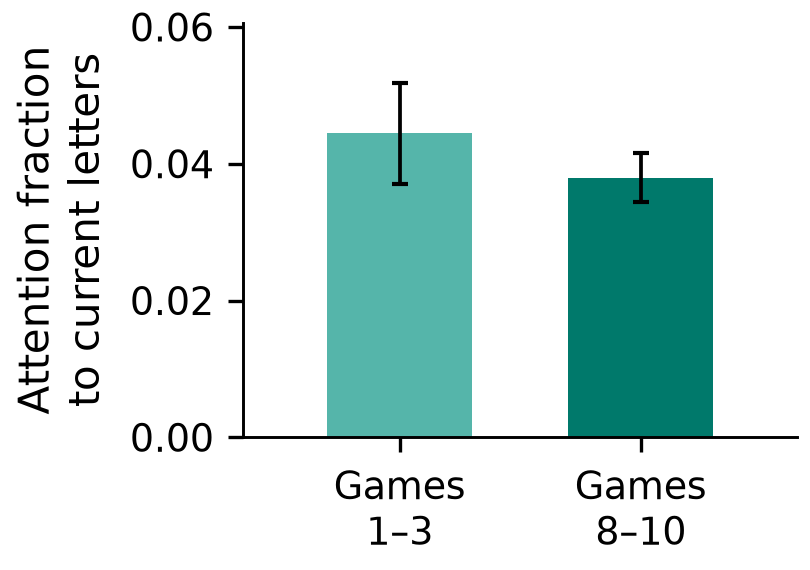}
\caption{\textbf{Less stimulus attention reaches current-game letters in final games.} Fraction of final-query attention reaching current-game letters, comparing initial games (Games 1--3) and final games (Games 8--10) in the two ICRL conditions. Attention is averaged over all heads in the five recorded layers at the final answer query, including games answered early. Trial means are averaged within each condition and game, then conditions and games are weighted equally within each model. Bars show means $\pm$ model SEM ($n=4$ models). This groups the same final-query measure shown by game in Figure~\ref{fig:icl-attention}f.}
\label{fig:icl-observed-attention}
\end{figure}

\clearpage
\subsection{Illustrative thinking trajectories in the other two models}
\begin{figure}[htbp]
\centering
\includegraphics[width=\linewidth]{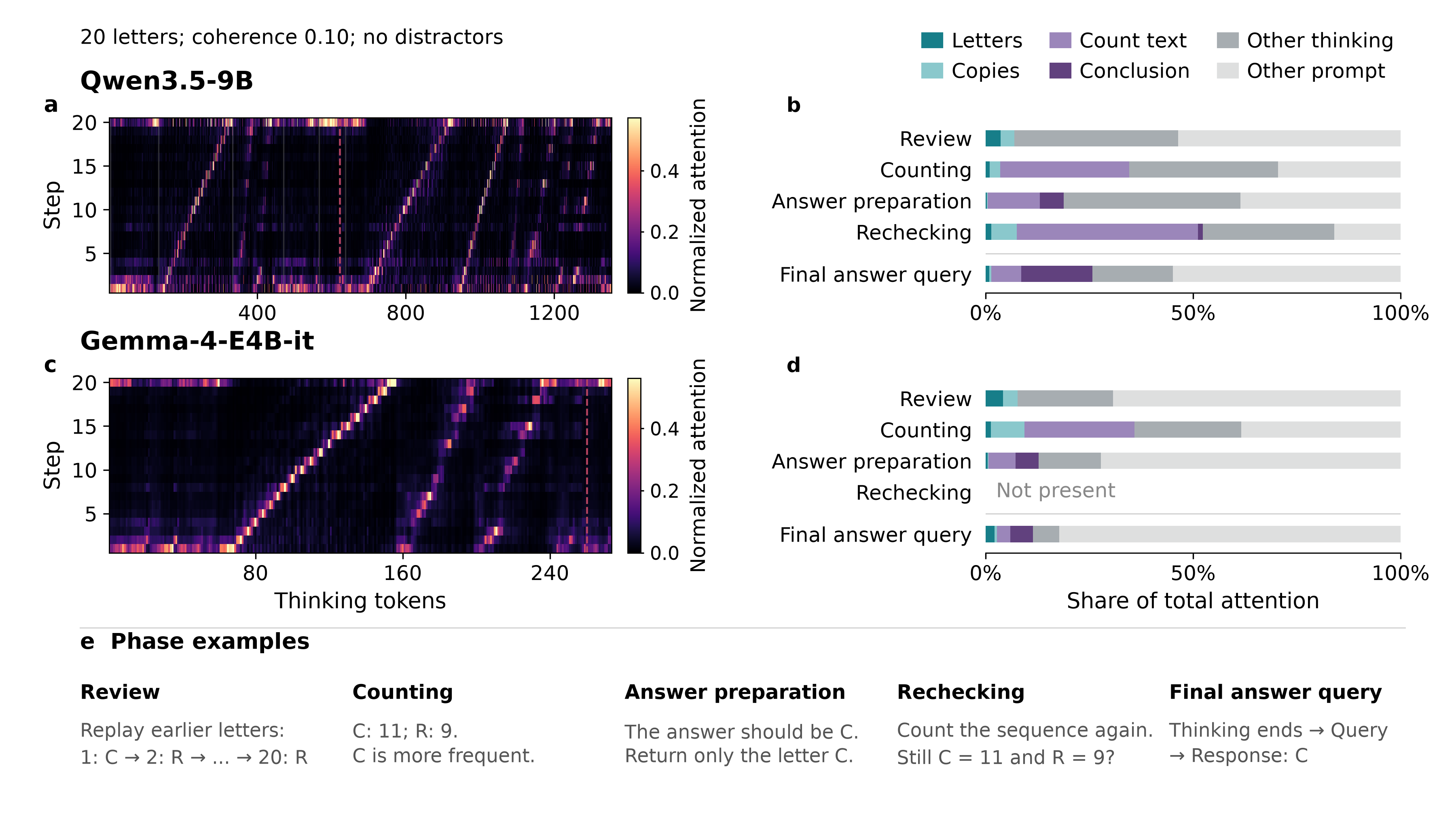}
\caption{\textbf{Selected Qwen3.5-9B and Gemma-4-E4B-it trajectories also revisit the letters during thinking.} Attention during thinking and across annotated reasoning phases in Qwen3.5-9B and Gemma-4-E4B-it. (a,c) Heatmaps in the same model order, normalized within each thinking-token column over original stimulus positions. Gray lines mark text-defined segment boundaries; dashed lines mark the first expressed preference, not an established internal decision time. Color limits are trial-specific. (b,d) Absolute attention shares across six token categories for queries in each reasoning phase (review, counting, answer preparation, and rechecking); the final answer query is shown separately. Segments with the same phase label are pooled by query count, so the displayed phase order need not be chronological. Missing phases are marked ``Not present.'' (e) Illustrative phase examples. Both trials use 20 letters, coherence 0.10, no distractors, and block 0, with the frozen opposite-fold single-game mass90 heads.}
\label{fig:thinking-other-models}
\end{figure}

Selected Qwen3.5-9B and Gemma-4-E4B-it trials under the same condition and block show the same progression through the letters as Figure~\ref{fig:thinking-heatmap} (Figure~\ref{fig:thinking-other-models}).

\end{document}